\documentclass{article}

\usepackage[english]{babel}
\usepackage{natbib}
\usepackage[letterpaper,top=2cm,bottom=2cm,left=3cm,right=3cm,marginparwidth=1.75cm]{geometry}
\usepackage{amsmath}
\usepackage{amssymb}
\usepackage{graphicx}
\usepackage[colorlinks=true, allcolors=blue]{hyperref}
\usepackage{threeparttable} 
\usepackage[table]{xcolor}
\newcommand{\headrow}{\rowcolor{black!20}}
\usepackage{float}
\usepackage{caption}

\title{AI-based mapping of the conservation status of orchid assemblages at global scale}
\author{Joaquim Estopinan, Maximilien Servajean, Pierre Bonnet, Alexis Joly, François Munoz}
\date{}

\begin{document}
\maketitle

\begin{abstract}
Although increasing threats on biodiversity are now widely recognised, there are no accurate global maps showing whether and where species assemblages are at risk. We hereby assess and map at kilometre resolution the conservation status of the iconic orchid family, and discuss the insights conveyed at multiple scales.
We introduce a new Deep Species Distribution Model trained on 1M occurrences of 14K orchid species to predict their assemblages at global scale and at kilometre resolution. We propose two main indicators of the conservation status of the assemblages: (i) the proportion of threatened species, and (ii) the status of the most threatened species in the assemblage. We show and analyze the variation of these indicators at World scale and in relation to currently protected areas in Sumatra island.
Global and interactive maps available online show the indicators of conservation status of orchid assemblages, with sharp spatial variations at all scales. The highest level of threat is found at Madagascar and the neighbouring islands. In Sumatra, we found good correspondence of protected areas with our indicators, but supplementing current IUCN assessments with status predictions results in alarming levels of species threat across the island.
Recent advances in deep learning enable reliable mapping of the conservation status of species assemblages on a global scale. As an umbrella taxon, orchid family provides a reference for identifying vulnerable ecosystems worldwide, and prioritising conservation actions both at international and local levels.
\end{abstract}

\section{Introduction}
\label{sec:intro}
Nearly a million species will face extinction in the coming decades \citep{diaz_pervasive_2019}, many of which having high value for medicine, food, materials, etc \citep{pollock_protecting_2020}. The Post-2020 Global Biodiversity Framework requires assessing current biodiversity state and quantifying conservation measures impacts \citep{nicholson2021scientific}. However, the distribution of many species is little known (Wallacean shortfall), and there is lack of comprehensive enough information on species conservation status \citep{schatz2009plants}. Land managers still need accurate indicators of species extinction risk that should be available both at a large scale (to allow comparisons between regions) and at a sufficiently fine spatial resolution. Recent automatic assessment of conservation status \citep{zizka_automated_nodate, borgelt_more_2022} have proved promising to complement the assessment based on informing IUCN criteria, which should help tackle the major objective of intensive prediction at broad taxonomic and spatial coverage.

Species distribution and richness patterns are complex, habitat and scale dependent, which entails that species conservation status must be assessed and acknowledged at multiple spatial scales and depending on habitat variation. According to \cite{whittaker_conservation_2005}, protected areas design based on species distribution and richness may be sensitive to spatial scale, and the conservation challenges must be addressed at both global scale and fine-resolution \citep{puglielli_macroecology_2023}. Here we perform (i) multiscale assessment of conservation status, based on (ii) high-resolution characterization of habitat properties, in the case of the emblematic orchid family.

Deep learning (DL) offers an unprecedented opportunity to characterize complex, scale-dependent relationships between species and their environment \citep{deneu_convolutional_2021}.
In addition, the ever-increasing volume of data stemming from citizen science observations on one hand, and from remote sensing characterization of environmental heterogeneity on the other hand, requires adapted DL workflows \citep{borowiec_deep_nodate}.
DL models can learn from complex effects and interactions between environmental predictors \citep{puglielli_macroecology_2023}, and \cite{cai_global_2022} have shown that DL can help to isolate relationships between biodiversity and ecological drivers.

Understanding how threatened species are distributed is a task that ecologists have been working on since the nineteenth century \citep{moret_humboldts_2019, gaston_spatial_1997}.
Yet there are few quantitative studies of the distribution of threatened species \citep{orme_global_2005}.
Successful attempts to design anthropogenic threat index at the regional scale \citep{paukert2011development} or even worldwide with the Human Footprint \citep{venter_global_2016} have lead the community to adopt this information as model predictor.
However, several major questions remain unsatisfactorily answered: how do anthropogenic and bioclimatic pressures relate to species environmental niches, at what scale and to what degree?
New studies in that regard consist in combining species IUCN status with known or predicted range of species and produce conservation priority maps \citep{han_integrated_2019, mair2021metric, verones_global_2022}.
Species included in these indices must have been previously assessed and their extinction risk status officially recognised by the IUCN.
However, as of 2022, only 7\% of the world's described species have an IUCN status (15\% for the world's known plants) \citep{barometer}.
Ultimately, there is a strong case to be made for including unassessed species in the design of spatial threat indicators.

In order to widen the currently narrow IUCN coverage, automatic classification methods have made a breakthrough. A major research avenue has emerged from this urgent task \citep{walker_caution_2020}. Two families of methods coexist: approaches that estimate IUCN criteria variables in advance to compare with official thresholds \citep{dauby_conr_2017, stevart_third_2019}, and models that directly predict IUCN status after being trained with predictors and already assessed species \citep{borgelt_more_2022, zizka_iucnn_2021, nic_lughadha_use_2019, gonzalez-del-pliego_phylogenetic_2019}. Methods in the first category are easier to interpret by construction. However, newer predictive models achieve impressive performance. Research is also exploring the use of SDMs to inform conservation status thanks to their niche modelling capabilities \citep{syfert_using_2014, breiner_including_2017}.

SDMs are correlative models learning from the association of species observations with environmental predictors \citep{elith_species_2009}. These statistical tools are now widely used and ongoing methodological work continue to improve their convergence and predictive power \citep{pollock_understanding_2014, powell-romero_improving_2022, lembrechts_incorporating_2019}.
Applications at all scales contribute to grasp diversity patterns and help to hold invasive species back \citep{botella_jointly_2021}, highlight biodiversity hotspots \citep{hamilton_increasing_2022} or orient Protected Areas (PAs) design \citep{guisan_predicting_2013}.
Deep-SDMs embrace deep learning vision architectures to leverage rare and critical environment spatial patterns \citep{deneu_convolutional_2021, leblanc2022species}.
Indeed, spatial and temporal \citep{estopinan_deep_2022} contexts were proven significant to model rare species niches and species-rich regions diversity.
These models capture the shared environmental preferences between multiple species and let information flow from the most common to the rarest species without corrupting their specific features \citep{botella2018species}.
Spatially Explicit Models (SEMs) integrate the location of observations as a predictor variable. While ecologists discourage its use when modelling species' environmental preferences, it has been shown to significantly improve prediction performance and influence conservation planning \citep{domisch_spatially_2019}. SEMs can incorporate local heterogeneities, creating positive feedbacks and allowing patterns to emerge at larger scales \citep{deangelis_spatially_2017}.

Our main contribution is to produce kilometre-scale extinction risk maps of species assemblages on a global scale.
A species assemblage is simply defined as \textit{members of a community that are phylogenetically related}, where a community is \textit{a collection of species that occur in the same place at the same time} \citep{fauth1996simplifying}.
To do this, we trained a deep-SDM model on 1M observations of 14K species distributed worldwide.
We then developed a novel method to estimate species assemblages. 
Coupled with the species' IUCN status, the assemblages are then characterised by extinction risk indicators.
Interactive maps are available online at \url{https://mapviewer.plantnet.org/?config=apps/store/orchid-status.xml#}. 
To our knowledge, this is the first realisation of SDM-derived spatial indicators at such resolution, taxonomic and geographic coverage.
Four levels of analysis are also discussed:
i) How is the extinction risk of orchid assemblages distributed at different scales?
ii) Which zones appear to contain the most threatened assemblages?
iii) Is there a correlation between the diversity of orchids in a country and the proportion of threatened species? and finally
iv) In Sumatra, how do our indicators relate to current PA implementation?

\section{Materials and methods}

\subsection{Taxonomic focus: the \textit{Orchidaceae} family}
The \textit{Orchidaceae} family is a perfect taxon to guide our research, both because of its inherent nature and because of its large data coverage \citep{cribb2003orchid}.
This uniquely diverse taxon comprises around 31,000 species, making it one of the largest flowering plant families \citep{POWO}.
Diversity and aesthetic appeal of orchids have made them the focus of attention for botanists and enthusiasts for decades. This has resulted in both a rich scientific literature \citep{cozzolino_orchid_2005, givnish_orchid_2016} and a wealth of observations: 8M raw GBIF observations, including 6.8M with coordinates \citep{gbif_orchids}.
Orchids are present on all continents and are flowering in a very wide range of altitudes and habitats.
This is a crucial aspect as our modelling approach aims to capture and project species preferences worldwide.
The threats they face - habitat destruction, climate change, pollution and intensive harvesting - make them singularly vulnerable.
Moreover orchids are a relevant indicator of the health of their environment \citep{newman_orchids_2009}.
This well-known and change-sensitive family can be used as a proxy to identify ecosystem conservation priorities \citep{yousefi_using_2020}.
Understanding threats, monitoring populations and distributions, and raising awareness are other key conservation objectives for the group \citep{wraith_orchid_2020}.
Orchids are widely used by international institutions as flagship species to lead and give visibility to the conservation debate \citep{cribb2003orchid}.
The challenge of orchid conservation cannot be tackled at the species level alone. Large-scale and broad approaches should necessarily complement studies carried out on emblematic species with a high risk of extinction \citep{fay_orchid_2018}.

\subsection{Species assemblage prediction model}

\subsubsection{Definition}
Our model for predicting species assemblages is derived from what is called \textit{set-valued prediction} (or  \textit{set-valued classification}) in the machine learning community \citep{mortier2021efficient, chzhen2021set}. The model is trained on presence-only (single-label) data, but is then used to predict a set of labels by thresholding the output categorical probabilities.
In more details, let us consider the following species assemblage prediction problem with $C$ distinct species.
The input set made of the predictive features associated to each occurrence location is denoted $\mathcal{X} = \{x_1, ..., x_n\}$.
The matching species label set is $\mathcal{Y} = \{1, . . . , C\}$.
The objective is to learn a species assemblage predictor on a training dataset composed exclusively of presence-only occurrences $(x_1, y_1), ... , (x_{n_{t}}, y_{n_{t}}) \in \mathcal{X} \times \mathcal{Y}$. The pairs $(x_i, y_i)$ are supposed to be independently sampled from a unknown probability measure $\mathbb{P}_{X,Y}$. This joint measure can be decomposed into the marginal distribution measure over $\mathcal{X}$, $\mathbb{P}_{X}$, and the conditional distribution of $y$ given an input $x$ denoted $\eta(x)= (\eta_1(x), . . . , \eta_C(x))$ and equal to
\[ \mathbf{\eta}_k(x)= \mathbb{P}_{X,Y}(Y=k|X=x) \] 
Then, the assemblage of species likely to be present conditionally to $x$ can be defined as:
\[S^*_{\lambda}(x):= \left\{k \in \mathcal{Y}: \mathbf{\eta}_k(x) \geq \lambda \right\} \] 
where $\lambda$ is a threshold on the conditional probability of species optimised to return precautionary assemblages (see next section on model validation).\\
In practice, the true conditional probability $\eta(x)$ is unknown and we assume we are given an estimator $\hat{\eta}(x)$ from which we can derive the following \textit{plug-in} estimator of the species assemblage:
\begin{equation}
S_{\lambda}(x):= \left\{k \in \mathcal{Y}: \hat{\eta}_k(x) > \lambda \right\}
    \label{equ:set_predict}
\end{equation}
One approach to get a good estimator $\hat{\eta}_k(x)$ of the conditional probability is to fit a model using the negative log-likelihood which is known to be a strictly proper loss \citep{gneiting2007strictly}, i.e. it is minimized only when the model predicts $\eta$. The negative log-likelihood loss is defined as:
\begin{equation}
    l_{log}(k,\hat{\eta})=-log \; \hat{\eta}_k(x)
    \label{equ:loss}
\end{equation}
In the context of deep learning, $\hat{\eta}(x)$ is typically chosen as a softmax function on top of a deep neural network $f_{\theta}(x):\mathcal{X}\rightarrow \mathbb{R}^C$ so that:
\[\hat{\eta}_k(x)=\frac{exp(f_{\theta}^k(x))}{\sum_j exp(f_{\theta}^j(x))}\]
where $\theta$ is the set of parameters of the neural network to be optimized by minimizing the loss function of equation \ref{equ:loss}.\\
Using this very common deep learning framework, it is possible to show that the species assemblage predictor $S_{\lambda}(x)$ of Equation \ref{equ:set_predict} is consistent \citep{lorieul2020uncertainty}, i.e. it tends towards the optimal set $S_{\lambda}^*(x)$ when the number of training samples increases. In other words, our species assemblage predictor is as simple as training a deep neural network with a \textit{cross-entropy} loss function on the presence-only samples and thresholding the output softmax probabilities to get the assemblage of predicted species.\\ 

Our backbone model is an adaptation of the Inception v3 \citep{szegedy_rethinking_2016}.
This convolutional neural network learn spatial patterns from two-dimensional predictors \citep{botella2018deep, deneu_convolutional_2021}.
A spatial block hold-out strategy is used to limit the effect of spatial autocorrelation in the data when evaluating the model \citep{roberts2017cross}. Blocks are defined in the spherical coordinate system according to a 0.025° grid (2.8 km square blocks at the equator).
The split of the training/validation/test spatial blocks (90\%/5\%/5\%) is stratified by region to ensure that all regions are represented within each set. We use the regions defined by the World geographical scheme for recording plant distribution (WGSRPD) level 2 \citep{brummitt2001world}.
Training is done on Jean Zay, a supercomputer from the Institute for Development and Resources in Intensive Scientific Computing (IDRIS).
A full description of the model architecture, dataset spatial split and training procedure can be found in supplementary information (SI) \ref{si:archi_training}.
Finally, the species assemblages are post-processed. i) Predictions outside the continents where species are known to occur (according to our oservation dataset) are removed, and ii) conditional probabilities associated with orchids are normalised, see SI \ref{si:oor}.

\subsection{Validation}
\label{ssec:val}

The species assemblage model is calibrated and assessed on the unseen occurrences from the validation spatial blocks (see dataset split in SI \ref{si:archi_training}).
The objective is to guarantee that the true species is included within the kept species assemblage.
This optimises recall rather than model precision.
It results in species assemblages that are potentially larger than in reality, and consequently in aggregated indicators at species level that are potentially overestimated but precautionary (see next section).\\
\noindent Our dataset is highly unbalanced in terms of the number of occurrences per species (see SI \ref{si:occ_distributions}).
It is therefore difficult to calibrate a specific threshold for many species. However, this would have been appropriate if we wanted to guarantee an error per species rather than per observation point.
The aim is indeed to reduce the marginal error of classification per observation (i.e. we want assemblages with little error on the species observed). The optimal solution is given by a common threshold per species \citep{fontana2023conformal}.

The threshold value $\lambda$ is then an important hyper-parameter of the method. Theoretically, we could consider that any species with a non-null conditional probability $\eta_k(x)$ is potentially present in the assemblage (i.e. by chosing $\lambda=0$). However, in practice, the estimator $\hat{\eta}_k(x)$ is never null even for the most unlikely species. Thus, it is required to adjust the value of $\lambda$ so that only the relevant species are returned in the assemblage. Therefore, we use a subset of the training dataset that is used only for this calibration step. It allows estimating the average error rate for a given value $\lambda$:
\[\mathcal{E}(S_\lambda) = \mathbb{P}_{X,Y}[Y \notin S_{\lambda}(X)]\]
by computing the percentage of samples $x_i$ in the calibration set for which the true observed species $y_i$ is not in $S_{\lambda}(x_i)$.

Finally, we can chose $\lambda$ so as to minimize the average species assemblage size $\mathbb{E}[|S_\lambda(X)|]$ - which is equivalent to maximize $\lambda$ - while guarantying that the average error rate is lower than an $\epsilon$ objective:
\begin{equation}
    \label{equ:lambda_optim}
    \begin{aligned}
        \mathop{\arg \min}\limits_{\lambda \in [0,1]}&\ \mathbb{E}[|S_\lambda(X)|]\\
        s.t.& \hspace{3pt} \mathcal{E}(S_\lambda) \leq \epsilon
    \end{aligned}
\ \Leftrightarrow\ 
    \begin{aligned}
      \max \limits_{\lambda \in [0,1]}&\ (\lambda)\\
        s.t.& \hspace{3pt} \mathcal{E}(S_\lambda) \leq \epsilon
    \end{aligned}
\end{equation}
This is equivalent to what is called conformal prediction in machine learning \citep{fontana2023conformal} and guarantees that the actual species is contained within the set with probability at least $1-\epsilon$. 

In practice, we choose $\epsilon = 0.03$ as explained in more details in the SI \ref{si:modelEval}.
We predict assemblages that have been validated to contain the initial species in 97\% (at the point level) and 80\% (at the species level) of cases.
The performance at the species level shows the robustness of our assemblages and the performance at the point level its validity in space.

\subsection{Conservation indices for species assemblages}

\subsubsection{Indices definition}
\label{sssec:I_def}
We define two indices characterizing the extinction risk of a predicted species assemblage, $\mathcal{I}_c \text{ and } \mathcal{I}_\mathcal{O}$.
They respectfully render the proportion of threatened species in the assemblage  and the most critical IUCN status in the assemblage. Let's break down their construction.

\paragraph{IUCN status notations}
Our indices partly rely on the extinction risk classification scheme from the IUCN Red List of threatened species, \url{https://www.iucnredlist.org/} \citep{mace_quantification_2008}.
IUCN categories are limited to \textit{Least Concerned} (LC), \textit{Near Threatened} (NT), \textit{Vulnerable} (VU), \textit{Endangered (EN)} and \textit{Critically Endangered} (CR).
We set the ensemble $E_{\text{status}}=$\\ $\{\text{LC}, \text{NT}, \text{VU}, \text{EN}, \text{CR}\}$ with the relation order LC$<$NT$<$VU$<$EN$<$CR.
Additionally, we introduce a general THREAT category corresponding to the union of VU, EN and CR categories.
We denote as $\varphi(y)$ the function that provides the extinction risk status of a species $y$.

\paragraph{Indicator $\mathcal{I}_\mathcal{O}(S)$: most critical status of the species in the assemblage}
For a given species assemblage $S$, our first indicator consists in taking on the most critical species extinction risk status. This is a concise and precautionary index. It aims at providing an information easy to understand and represent. Here is its formal definition:
\begin{equation}
    \begin{aligned}
        & \mathcal{I}_\mathcal{O}: && \mathcal{P}(\mathcal{Y}) && \mapsto     && E_{\text{status}}\\
        &                          &&     \hspace{5pt}S             && \rightarrow && \max_{s_j \in S}\left\{ \varphi(s_j)\right\}
    \end{aligned}
\end{equation}

\paragraph{Indicator $\mathcal{I}_c(S)$: proportion of species in the assemblage with a given status}
Our second indicator $\mathcal{I}_c(S)$ measures the proportion of species from a given category $c$ in an assemblage $S$.
Let us consider a species assemblage with its associated probability distribution $(S,\eta)$.
$\mathcal{I}_c$ is defined as the proportion of species with status $c$ in $S$, with the species being weighted by their relative probability of presence $\eta$ (see Equation \ref{equ:I_T}).
The proportion of critically endangered species is for instance denoted $\mathcal{I}_\mathcal{\text{CR}}(S)$. And so on for the four other IUCN status in $E_{\text{status}}$ and the overall THREAT category. 
\begin{equation}
\label{equ:I_T}
\begin{aligned}
&\mathcal{I}_c : && \mathcal{P}(\mathcal{Y})\times\mathbb{R}^C && \mapsto && \mathbb{R}^{[0,1]}\\
&                            && \hspace{15pt}(S, \eta)      && \rightarrow && \sum\limits_{j \in \varphi^{-1}(c)} \eta_j
\end{aligned}
\end{equation}

\paragraph{The Shannon index $\mathcal{I}_\mathcal{H}(S)$}
The Shannon index is one of the most popular measures of biodiversity. It originates from the famous communication theory \citep{shannon1948mathematical}, but was adopted in ecology as early as 1955 \citep{ricotta2005through}. Denoted $\mathcal{I}_\mathcal{H}$, this metric evaluates the quantity of information of a set.
Both the set richness (number of distinct classes) and evenness (classes ratio) influence the index \citep{marcon_mesures_2015}.
Let $(S,\eta)$ be a species assemblage, with $\eta$ its associated conditional probability distribution:
\begin{equation}
    \label{equ:shannon}
    \mathcal{I}_\mathcal{H}(S) = - \sum\limits_{l \in S} \eta_l \cdot \log(\eta_l)
\end{equation}

\subsubsection{Missing status completion} 
Only 889 of our 14,129 orchid species have an official IUCN status in 2021, i.e. 6.3\%. It therefore seems unreasonable to ignore all unassessed species in our indicator calculation. We decide to supplement the status information with an automatic preliminary assessment method from the literature called IUCNN \citep{zizka_iucnn_2021}.
The distributions of the IUCN-assessed and predicted IUCN status are shown in Figure \ref{si:status_distrs}.
Both indicators can then be computed considering only IUCN-assessed species or the entire species assemblage. By default, the indicators are on all the orchid species from our assemblage, i.e. considering both known IUCN status and predicted IUCN status. When they are restrained on the IUCN-assessed species only, the indicators are denoted with an \textit{IUCN} superscript: $\mathcal{I}^{\text{IUCN}}$.

\subsection{High-resolution maps construction}

\subsubsection{Global grid design}
The aim now is to create a global grid to support our spatial indicators. This is done in two steps.
First, we create a regular grid covering all longitudes and latitudes. We sample the longitude range [-180°,180°] and the latitude range [-90°,90°] at 30-second intervals. One second equals 1/3600 degrees, hence $r=30/3600$ degrees.
Let $\mathcal{M} = \{-180, -180+r,..., 180-r,180\}$ and $\mathcal{N}$ be its latitudinal counterpart.
The grid support is then obtained by crossing the two sampled axes $\mathcal{M}\times\mathcal{N}$.
Secondly, we spatially intersect the grid with the land areas of the world. We are indeed only interested in terrestrial regions. The geometry used is the \textit{Esri} grid of world country boundaries \citep{arcgis}. The intersection contains 221M points. Finally, predictive features are assigned to each land grid position. This results in $\mathcal{G} = \{x_{m,n} \hspace{2pt}|\hspace{2pt} m,n \in \mathcal{M}\times\mathcal{N}\}$.

\subsubsection{Maps definition and construction}
Maps are constructed in two steps: First, the species assemblages associated to each $\mathcal{G}$ grid point are predicted by batch with our model:
$\hat{\mathcal{S}}_\lambda(\mathcal{G})$.
Second, the spatial indices defined in section \ref{sssec:I_def} are computed on the predicted assemblages:
$\mathcal{I}(\mathcal{G}) = \{\mathcal{I}(S) \hspace{2pt}|\hspace{2pt} S \in \hat{\mathcal{S}}_\lambda(\mathcal{G})\}$.
This set of indicators $\{ \mathcal{I}(\mathcal{G}) \}$ constitute our global and kilometre-scale maps (\textit{reminder:} by default all orchid species are considered and predicted IUCN status thus employed). Within worldwide predicted species assemblages:
\begin{itemize}
    \item $\mathcal{I}_{\mathcal{O}}(\mathcal{G})$ highlights the most critical IUCN status
    \item $\mathcal{I}_{c}(\mathcal{G})$ represents the proportion of species with IUCN status $c$ (five maps)
    \item $\mathcal{I}_{\text{THREAT}}(\mathcal{G})$ maps the proportion of threatened species
    \item $\mathcal{I}_{\mathcal{H}}(\mathcal{G})$ draws the global patterns of predicted orchid diversity.
\end{itemize}
Details on predictions batch processing and on the website solution are available in \ref{si:maps_batch_and_access}.

\subsection{Zonal statistics}   
Spatial analysis can necesit aggregated regional indicators.
With a kilometre scale resolution, $\mathcal{I}_\mathcal{O}$ and $\mathcal{I}_c$ can be dissolved at different organization levels. 
Municipalities, protected areas, states or biodiversity units: the choice depends on the application. To illustrate this method at the global scale, we aggregate our indicators at the WGSRPD level 3.
It corresponds to \textit{botanical countries} which can ignore political borders \citep{brummitt2001world}.
We selected countries of at least 2,000~km² to highlight large area priorities (65 countries out of 369 removed).

\subsubsection{Region spatial coverage of the most critical IUCN status}
This measure is based on $\mathcal{I}_\mathcal{O}$, the spatial indicator of the most critical IUCN status in the species assemblage. In a given region $r$, areas with distinct worst IUCN status coexist. Focusing on a given status $c$, its spatial coverage proportion in $r$ is denoted $\text{Area}_\%[\mathcal{I}_\mathcal{O}](r,c)$. By default, this variable is computed on the entire species assemblage. Nonetheless, it can also be expressed considering only IUCN-assessed species.

\subsubsection{Region average proportions}
Second zonal statistic consists in taking $\mathcal{I}_c$ average for a given region $r$ and status $c$. It represents region's average proportion of species with $c$ as IUCN status and is written down $\mu[\mathcal{I}_c](r,c)$. 
The entire species assemblage is taken into account. Such statistic allows direct comparison between arbitrary zones.
For the sake of simplicity, square brackets precising the spatial indicator can be dropped in both zonal statistics.

\subsection{Data}
\subsubsection{Orchid occurrences}

The orchid occurrence dataset comes from \cite{zizka_automated_nodate}, whose authors queried GBIF in August 2019. This dataset has the advantage of being both global and already geographically/taxonomically curated. Nearly 1 million occurrences of 14,129 different species were used to build our model (999,258 observations after duplicate checking). The average number of observations per species is 70, while the median is 4. 25\% of species have more than 13 species. Date distribution summary statistics are $\text{min}=1901, \text{Q1}=1982, \text{med}=1997, \text{Q3}=2010 \text{ and max}=2019$.
The cumulative number of occurrences per species, the distribution of observation dates, the distribution of georeferencing uncertainty, the observation map and the species richness maps are all available in SI \ref{si:occ_distributions}.

\subsubsection{Predictive features} 
A large environmental context around each observation is collected and provided to the model: 64 x 64 2D tensors sampled at the kilometre-scale resolution and centred on the observation.
Predictors include WorldClim2 bioclimatic variables, Soilgrids pedological variables, human footprint rasters, terrestrial ecoregions of the world and the observation location, see SI \ref{si:predictors} for details.
Examples of input are shown in Figure \ref{si:input_exs} and the full list of predictors is given in Table \ref{tab:features_table}.

\section{Results}

\subsection{$\mathcal{I}_\mathcal{O}$ indicator: most critical status of the species in the assemblage}

\subsubsection{Global patterns}

\begin{figure}[hp]
\centering
\includegraphics[width=\textwidth]{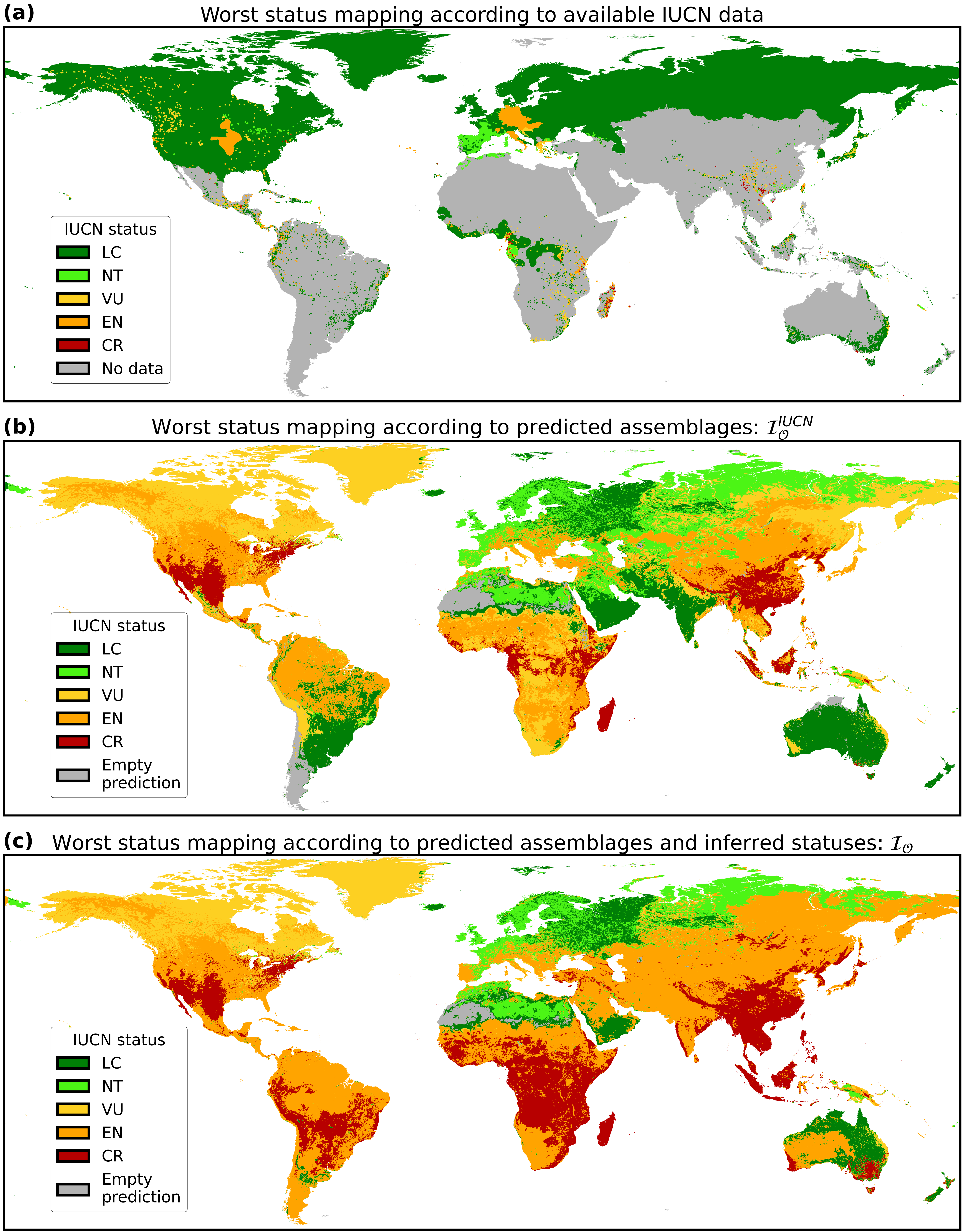}
\caption{Global comparison of the most critical IUCN status indicator according to three methods.
(a) represents the IUCN information on our dataset: observations and available spatial data (polygons and points from
\textit{https://www.iucnredlist.org/resources/spatial-data-download}) taken together.
Spatial data is available for only 167 IUCN-assessed orchids from our dataset, i.e. 1.2\% of all species.
(b) is the result of our species assemblage prediction model coloured by the most critical known IUCN status whereas (c) includes predicted IUCN status too in the indicator calculation.
[Figure maps are under-sampled, see the website for full-resolution]}\label{fig:rasters_comparison}
\end{figure}

Considering the worst status of a species assemblage, Figure \ref{fig:rasters_comparison} compares (a) currently available IUCN information with (b,c) our model results $\mathcal{I}_\mathcal{O}^{\text{IUCN}}$ and $\mathcal{I}_\mathcal{O}$.
IUCN species range data are still very scarce (only 1.2\% of species in our dataset have IUCN ranges) and of variable quality: some species have raw model outputs as official IUCN range maps whereas others will have tailored expert-designed maps.
Our species assemblage model combined with known IUCN status results in a consistent and contrasted map Fig. \ref{fig:rasters_comparison}b.

\noindent Predictions in tropical Africa, East and South-East Asia and North America include CR species assessed by the IUCN.
The presence of CR species in North America may be surprising at first, but given that i) this continent is comparatively well assessed and ii) this indicator is both sensitive and precautionary (only one species is sufficient to reach the CR category), it is reasonable.
No CR species are predicted in South America if only known IUCN status are considered.
However, when predicted IUCN status are included on Fig. \ref{fig:rasters_comparison}c, the value of $\mathcal{I}_\mathcal{O}$ across South America is drastically different. Indeed, EN and CR species predictions lead the indicator to change to higher categories of risk.
According to our model taking into account predicted IUCN status, Brazil and the Andes are for instance hosts to CR-estimated species on a large part of the territory.
On Figure \ref{fig:rasters_comparison}c, new global patterns are highlighted. These include India and temperate Asia presenting EN species, the Western Ghats and Southeast Asia hosting CR species, and Portugal, western Spain and the French Landes turning orange due to the prediction of EN species.
Overall, the differences are more pronounced in the southern hemisphere than in the northern hemisphere.
This illustrates the fact that IUCN assessments are biased towards northern countries and that large assessment gaps remain.

\subsubsection{Country-level analysis}

\begin{table}[tb]
\centering
\begin{threeparttable}
\caption{
Top-15 countries with the largest share of their area covered by CR (\textit{left}) or EN \textit{(right)} as most critical IUCN status.
}\label{tab:Area}

\begin{tabular}{llrlr}
\headrow
{} & \multicolumn{2}{l}{\textbf{CR}} & \multicolumn{2}{l}{\textbf{EN}} \\
\headrow
{} & \textbf{B. country} & \textbf{Area}$_\%$ & \textbf{B. country} & \textbf{Area}$_\%$ \\
\textbf{1 } &          Eq. Guinea &             100.00 &             Jamaica &             100.00 \\
\textbf{2 } &             Réunion &             100.00 &      Dominican R. &             100.00 \\
\textbf{3 } &           Mauritius &             100.00 &               Haiti &              99.95 \\
\textbf{4 } &          Madagascar &              99.76 &                Cuba &              99.86 \\
\textbf{5 } &             Comoros &              99.60 &         Afghanistan &              99.74 \\
\textbf{6 } &                Laos &              99.38 &       French Guiana &              99.65 \\
\textbf{7 } &         Connecticut &              98.71 &              Guyana &              99.45 \\
\textbf{8 } &             Vietnam &              98.59 &             Surinam &              99.29 \\
\textbf{9 } &            Rhode I. &              98.49 &          Costa Rica &              99.15 \\
\textbf{10} &            Cambodia &              98.26 &            Portugal &              99.02 \\
\textbf{11} &                Jawa &              97.93 &               Corse &              98.98 \\
\textbf{12} &       Massachus. &              97.25 &        Tadzhikistan &              98.79 \\
\textbf{13} &          E Himalaya &              97.07 &         Puerto Rico &              98.71 \\
\textbf{14} &            Thailand &              96.99 &        Windward Is. &              98.64 \\
\textbf{15} &            Sumatra &              96.93 &           Galápagos &              98.50 \\
\hline  
\end{tabular}

\begin{tablenotes}
\item B. country, botanical country (WGSRPD level 3).
\end{tablenotes}

\end{threeparttable}
\end{table}

Table \ref{tab:Area} shows the botanical countries with the largest $\mathcal{I}_\mathcal{O}$ coverage as CR or as EN.
There are many islands in this ranking.
All top fifteen countries are almost completely covered by only one status.
See supplementary information T3 for the full table.
High on the $\text{Area}_{\%}(\text{CR})$ ranking are Equatorial Guinea, Réunion, Mauritius, Madagascar, Comoros and Laos.
CR species are present throughout these countries.
By construction, countries with a high CR coverage status cannot also have a high EN coverage.
Therefore, countries with high $\text{Area}_{\%}(\text{EN})$ are different from the first column.
European territories such as Corse or Portugal appear in the ranking and Caribbean islands are well represented.

\subsection{$\mathcal{I}_c$ indicator: proportion of species in the assemblage with a given status}

\subsubsection{Global patterns}

\begin{figure}[hp]
\centering
\includegraphics[height=0.95\textheight]{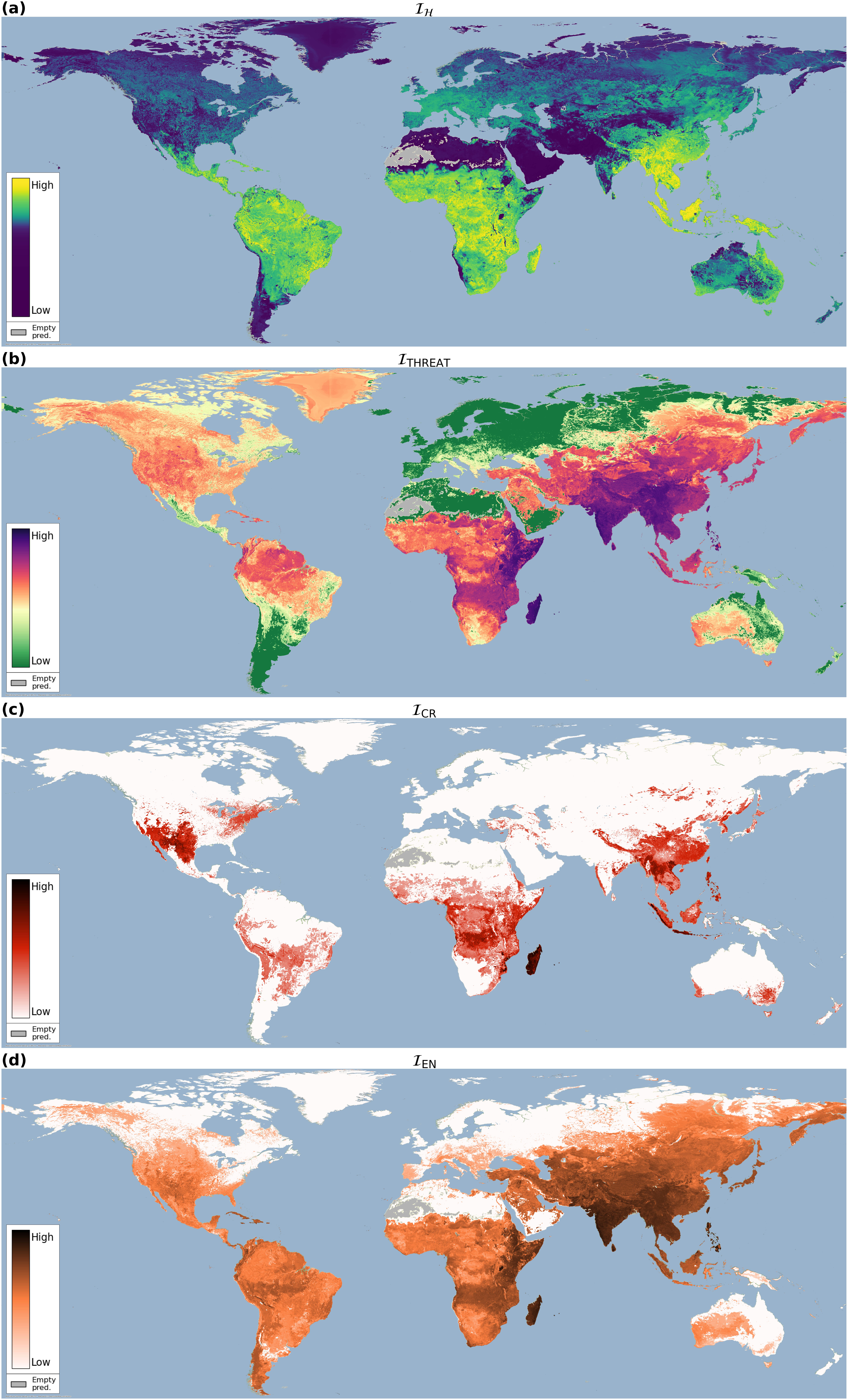}
\caption{Four indicators based on species assemblage predictions. (a) $\mathcal{I}_{\mathcal{H}}$ the Shannon index, (b) $\mathcal{I}_{\text{THREAT}}$ the weighted proportion of threatened species, (c) and (d) the weighted proportions of respectively CR species $\mathcal{I}_{\text{CR}}$ and EN species $\mathcal{I}_{\text{EN}}$.
[Figure maps are under-sampled, see the website for full-resolution]}\label{fig:Ic}
\end{figure}

Figure \ref{fig:Ic}a shows the Shannon index calculated on our species assemblage predictions (full resolution on the website).
As expected, the tropics appear to contain the richest areas.
This map can be read in parallel with the Figure \ref{si:richness_map}: the species richness map of our occurrence dataset stratified by botanical country (WGSRPD level 3).
The resolution gain is clear.
Moreover, some biases in the initial observations set explain $\mathcal{I}_{\mathcal{H}}$ patterns.
Colombia orchid richness, estimated for instance at 4,327 species according to \textit{World Plants} \citep{WP}, is for instance under-represented within our occurrence set with only 1,375 species.
Global orchid diversity patterns can also be appreciated in relation to the three following maps, which reflect the extinction risk of the predicted species assemblages.

\noindent High proportions of threatened species appear in East Africa, South and Southeast Asia on Figure \ref{fig:Ic}b $\mathcal{I}_{\text{THREAT}}$.
The Sahel also has a particularly high proportion of threatened species.
Orchids in central North America also appear to have relatively high rates of threatened species, given the low  observed and predicted diversity in this region.
The threat levels in the Amazon Basin are high. However, compared to East Africa or tropical Asia, they are not as high as the region's impressive orchid richness would suggest. This result is quantified on the scatter plot Figure \ref{fig:scatterplot}. High diversity does not necessarily imply high threat levels.

\noindent On Figure \ref{fig:rasters_comparison}c map (proportion of CR species), the first striking element is certainly the strong emphasis on Madagascar.
The patterns in the Himalayan belt, Indonesia and Southeast Asia are both more contrasted and appear more localised than on the $\mathcal{I}_{\text{THREAT}}$ (b) map.
In northern Mexico and the southwestern United States of America, high levels of CR species are appealing and contrasting with the Shannon index.
In South America, our model predicts relatively high levels of CR species along the Andes, in Bolivia, Paraguay and southern Brazil.
If we compare $\mathcal{I}_{\text{CR}}$ with $\mathcal{I}_{\text{CR}}^{\text{IUCN}}$ (see website), we can see that the presence of CR species in South America is almost entirely due to predictions whose IUCN status has been automatically classified.

\noindent Finally, $\mathcal{I}_{\text{EN}}$ levels (Fig. \ref{fig:Ic}d) are important throughout sub-Saharan Africa, Central and South America, South and Southeast Asia.
The patterns observed here are closer to $\mathcal{I}_{\text{THREAT}}$ than $\mathcal{I}_{\text{CR}}$.
With these maps we can better understand how the patterns of $\mathcal{I}_{CR}$, $\mathcal{I}_{EN}$ and $\mathcal{I}_{VU}$ indicators combine to produce the $\mathcal{I}_{\text{THREAT}}$ map.

\subsubsection{Country-level analysis}

\begin{table}[tb]
\vspace{-0.25cm}
\centering
\begin{threeparttable}
\caption{
Top-15 average status proportions per botanical country. \textit{From left to to right}: threatened species all taken together (THREAT), Critically endangered species (CR) and Endangered species (EN).
In average, 60\% of the predicted species in Madagascar are threatened by extinction (63\% in Réunion island).
}\label{tab:avg_prop}

\begin{tabular}{llrlrlr}
\headrow
{} & \multicolumn{2}{l}{\textbf{THREAT}} & \multicolumn{2}{l}{\textbf{CR}} & \multicolumn{2}{l}{\textbf{EN}} \\
\headrow
{} & \textbf{B. country} & $\mu[\mathcal{I}_c]$ & \textbf{B. country} & $\mu[\mathcal{I}_c]$ & \textbf{B. country} & $\mu[\mathcal{I}_c]$ \\
\textbf{1 } &             Réunion &                           0.63 &             Réunion &                           0.15 &             Réunion &                           0.44 \\
\textbf{2 } &          Madagascar &                           0.60 &          Madagascar &                           0.12 &          Madagascar &                           0.39 \\
\textbf{3 } &           Mauritius &                           0.55 &           Mauritius &                           0.10 &           Mauritius &                           0.38 \\
\textbf{4 } &             Comoros &                           0.48 &             Comoros &                           0.10 &               India &                           0.36 \\
\textbf{5 } &               Kenya &                           0.46 &                Jawa &                           0.07 &         Philippines &                           0.35 \\
\textbf{6 } &             Myanmar &                           0.45 &            Sumatra &                           0.04 &              Taiwan &                           0.34 \\
\textbf{7 } &               Nepal &                           0.45 &              Azores &                           0.03 &             Myanmar &                           0.33 \\
\textbf{8 } &          E Himalaya &                           0.44 &         Philippines &                           0.03 &           Sri Lanka &                           0.33 \\
\textbf{9 } &             Somalia &                           0.44 &             Vietnam &                           0.03 &          E Himalaya &                           0.33 \\
\textbf{10} &               India &                           0.44 &                Laos &                           0.03 &               Nepal &                           0.33 \\
\textbf{11} &                Laos &                           0.43 &             Arizona &                           0.03 &                Laos &                           0.32 \\
\textbf{12} &               Assam &                           0.43 &          New Mexico &                           0.03 &               Assam &                           0.32 \\
\textbf{13} &            China SC &                           0.42 &             Myanmar &                           0.03 &             Comoros &                           0.30 \\
\textbf{14} &          W Himalaya &                           0.40 &          Mozambique &                           0.03 &            Thailand &                           0.30 \\
\textbf{15} &              Taiwan &                           0.40 &    Lesser Sunda Is. &                           0.03 &            Cambodia &                           0.29 \\
\hline
\end{tabular}


\end{threeparttable}
\end{table}

In Table \ref{tab:avg_prop}, the top three botanical countries with the highest average proportion of threatened species, species classified as CR and species classified as EN are common: Réunion Island, Madagascar and Mauritius Island.
Overall, 60\% of the species predicted for Madagascar are threatened with extinction.
All $\mu[\mathcal{I}_{\text{THREAT}}]$ top fifteen countries have an overall predicted proportion of threatened species greater than or equal to 40\%.
Again, the three columns are dominated by East African and tropical Asian countries.
See supplementary information T3 for the full table.

\begin{figure}[tb]
\centering
\includegraphics[width=\textwidth]{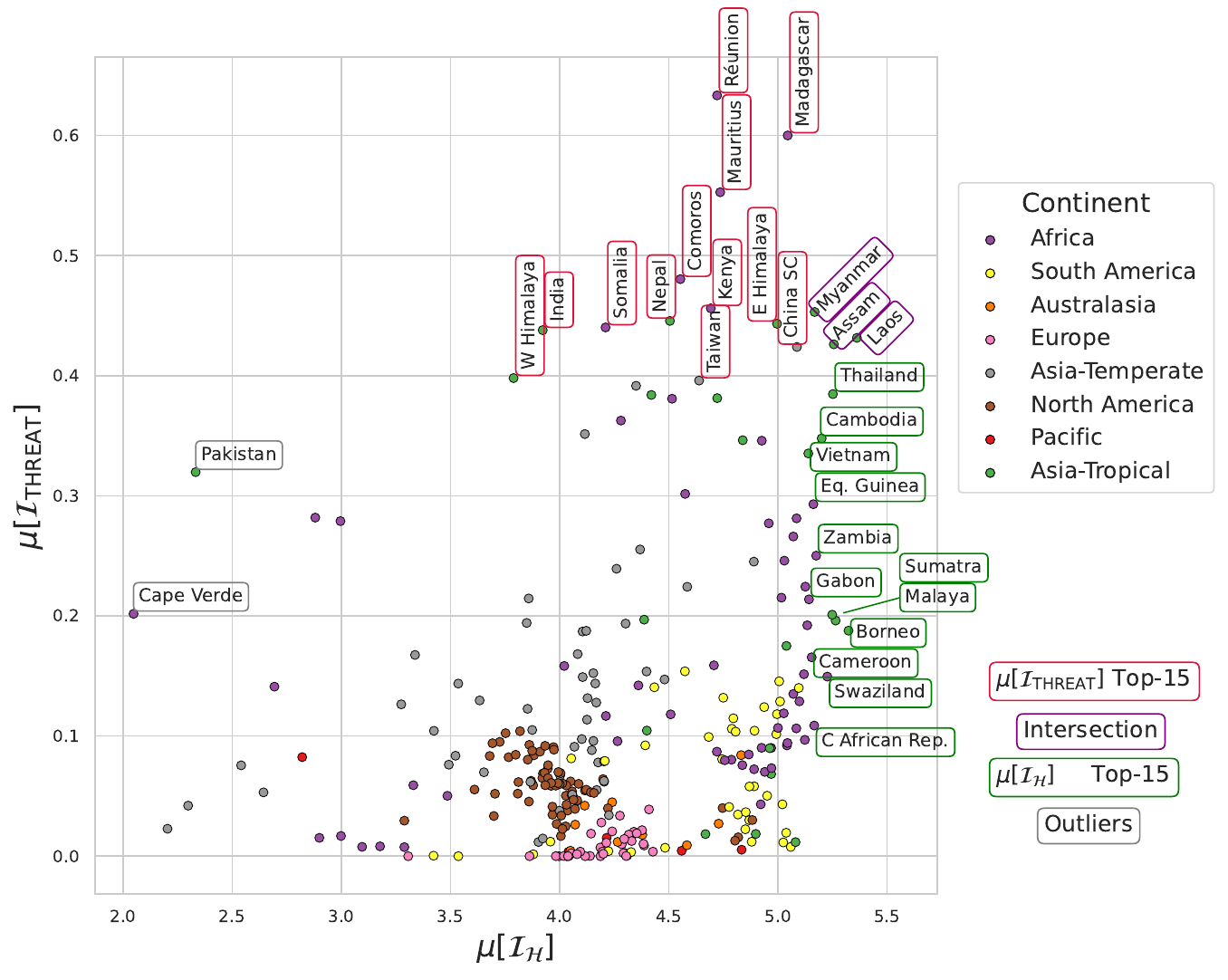}
\caption{Average proportion of species predicted as threatened  by botanical country (WGSRPD level 3) \textit{versus} average Shannon index. Countries are coloured in function of their continent (WGSRPD level 1) and top-15 countries of both variables are highlighted. Myanmar, Assam and Laos are the only three regions in the top-15 intersection whereas Pakistan and Cape Verde show especially high threatened species proportions with low diversity indices. }\label{fig:scatterplot}
\end{figure}

The scatterplot Figure \ref{fig:scatterplot} tests the relation between the average rate of threatened species and the Shannon index at the level of botanical countries.
The Spearman $\rho$ value is $0.29$ ($p = 2.5\mathrm{e}{-7}$), indicating a positive but relatively law global correlation. 
The colour code, indexed by continent, reveals different patterns per continent.
North American (brown) and European (pink) countries are clearly clustered on the graph, with a medium diversity index and low threat levels on average.
The top fifteen $\mu[\mathcal{I}_{\text{THREAT}}]$ countries (table \ref{tab:avg_prop} first column) are this time marked with red borders.
The top fifteen $\mu[\mathcal{I}_{\mathcal{H}}]$ are framed in green and the intersection includes Myanmar, Assam and Laos.
African (purple), Asian temperate (grey) and Asian tropical (green) countries present more variation in this graph and represent the extremes.
The South American countries (yellow) at the bottom right of the graph confirm the observation made with Figure \ref{fig:Ic}: this continent is highly diverse with relatively low levels of threat to its species assemblages.
A Venn diagram crossing $\mu[\mathcal{I}_{\mathcal{H}}]$ and $\mu[\mathcal{I}_{\text{THREAT}}]$ top-30 countries plus the Spearman correlations per continent are available at Figure \ref{si:venn_rho_pC}.

\paragraph{Sumatra case study}  

\begin{figure}[tb]
\centering
\includegraphics[width=\textwidth]{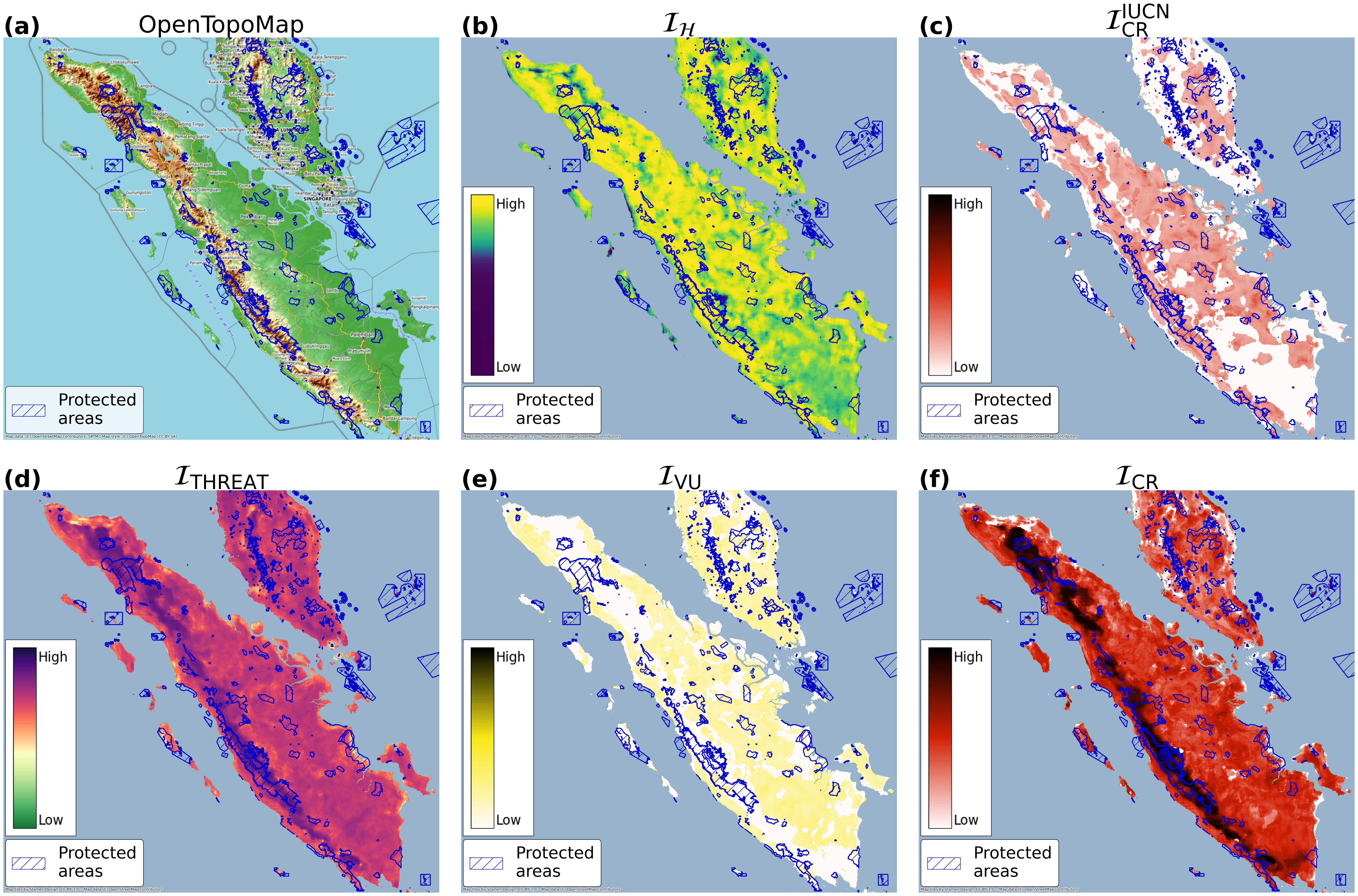}
\caption{Five indicators of species assemblage extinction risk applied on Sumatra island. Elevation is also provided and protected areas are hashed in blue (downloaded from \textit{https://www.protectedplanet.net/en}). (a) elevation map, (b) Shannon index, (c) proportion of IUCN-assessed CR species in the predicted species assemblages. On the second line, species proportion of: (d) threatened species, (e) VU species only, and (f) CR species only (all statuses combined).
[Maps in figures are under-sampled, see the website for full-resolution] }\label{fig:Sumatra}
\end{figure}

On the western side of Sumatra, the Barisan Mountains form a sharp relief (see Figure \ref{fig:Sumatra}a). The \textit{elevational diversity gradient} theory would suggest that species richness is particularly high along the mountainous area.
However, according to the $\mathcal{I}_{\mathcal{H}}$ indicator on (b), the predicted orchid diversity appears to be fairly constant across the island.
Considering only the known IUCN assessments, the presence of CR species (c) is not clearly correlated with the mountain range.
In addition, there are areas where no CR species are predicted, for example in the northern and southern regions of the island.
When the predicted IUCN status are included in the indicator calculation with $\mathcal{I}_{\text{CR}}$ on (f) map, high proportions of CR species are predicted across the island. There is a sharp pattern following the Barisan Mountains.
By construction, a similar trend is drawn on the (d) map representing $\mathcal{I}_{\text{THREAT}}$.
Such a difference between $\mathcal{I}_{\text{CR}}$ and $\mathcal{I}_{\text{CR}}^{\text{IUCN}}$ at the regional scale confirms the need to include automatic IUCN assessments when designing extinction risk indicators.
Finally, $\mathcal{I}_{\text{VU}}$ on Fig. \ref{fig:Ic}e map indicates the likely presence of VU species inhabiting the lower elevations of the islands.

Protected areas cover 12.7\% of the island of Sumatra.
Three national parks on the spine of the Barisan Mountains were inscribed on UNESCO's World Heritage List in 2004, forming the Tropical Rainforest Heritage of Sumatra.
They are the three largest protected areas on the island.
From north to south: Gunung Leuser National Park, Kerinci Seblat National Park and Bukit Barisan Selatan National Park.
Since 2011, these parks have been placed on a Danger List to help combat numerous threats, including poaching, illegal logging and agricultural encroachment.

Let's look at the zonal statistics for PAs.
We calculate the ratio of two indicators, both averaged across PAs: i) the proportion of \textit{all} CR species (known IUCN status + predicted status combined) and ii) the proportion of \textit{IUCN-assessed} CR species: $\frac{\mu[\mathcal{I}_{\text{CR}}]}{\mu[\mathcal{I}_{\text{CR}}^{\text{IUCN}}]}(\text{PAs}) = 3.1$.
This ratio is even greater when all threatened species are considered together:
$\frac{\mu[\mathcal{I}_{\text{THREAT}}]}{\mu[\mathcal{I}_{\text{THREAT}}^{\text{IUCN}}]}(\text{PAs}) = 7.1$
The level of threat in Sumatra's PAs is then significantly higher than the IUCN information alone would suggest.
Now let's compare the average CR proportion inside \textit{versus} outside PAs:
$\mu[\mathcal{I}_{\text{CR}}](\text{PAs}) = 0.108$ and  $\mu[\mathcal{I}_{\text{CR}}](\overline{\text{PAs}}) = 0.036$. Thus the average proportion of CR species is 3 times higher in PAs than outside PAs.
The current design of PAs therefore seems to well match habitats hosting particularly threatened orchids.
However, looking closely at the map reveals that many areas with a specially high proportion of CR species are still outside PAs, so that the ratio could be consistently improved.
With IUCN-assessed species only, the average proportion of CR species in PAs is 3.4\%.
It is similar to the proportion of CR species \textit{outside} PAs with the completed Red List.
Again, enriching the current IUCN information within our method changes the narrative on PA efficiency.

\section{Discussion}

\subsection{Modelling choices and considerations on covariates}
Our species assemblage predictor has theoretical guarantees that we have validated on a previously unseen observation set (see SI \ref{si:modelEval}). However, some bias in the input data could prejudice its predictions. Unlike some methods, it has the advantage of not being biased by the heterogeneous sampling effort. Indeed, it depends only on the conditional probability $\mathbb{P}_{X,Y}(Y=k|X=x)$ and not on the marginal distribution $\mathbb{P}_{X}$. Nonetheless, it is impacted by species detection bias, i.e. by the fact that some species might be observed more than others conditionally to a given $x$. Largely under-observed species, in particular, may be excluded from the predicted assemblage. Conversely, some over-observed species could be predicted at locations where they are not present. In future work, it would be interesting to study the impact of this type of bias on the assemblage-level indicators introduced in this paper.
Further considerations on the model (on the trade-off between model generalisation and over-prediction, on the difficulty of measuring the precision of the model) are carefully detailed in \ref{si:model_data_consid}.

Nature’s myriad of elements are interfaced to produce heterogeneous patterns of diversity, unpredictable at a given point, but statistically structured.
Measuring some of these factors and feeding them into our model will hopefully allow us to capture biodiversity shapes.
However, it is essential to remember that no single mechanism fully explains a given pattern, that inter-scale dependencies and local historical events strongly influence biodiversity, and that no pattern is exempt from variation and exceptions \citep{gaston_global_2000}.
Other ecological variables contain valuable information influencing the distribution of orchids.
They have not been included because of the currently limited spatial and taxonomic coverage or for practical reasons.
Remote sensing is a natural perspective for improvement \citep{he_will_2015, gillespie_image_2022}.
The inclusion of biological and functional traits of orchids is another exciting perspective \citep{puglielli_macroecology_2023, weigelt_gift_2020, bourhis2023explainable}, as well as mycorrhizal fungi or pollinator distribution \citep{mccormick_mycorrhizal_2018}.

We believe that predictors of large spatial patterns may play a significant role in the regional diversity of orchids, and that the computer vision model can learn such information.
The model’s strength is to rely on the best possible input set and exploit complex interactions in order to be as predictive as possible.
The trade-off is interpretability, but the AI community is investing heavily in this area and our understanding is getting finer \citep{linardatos_explainable_2021}.
For example, deep-SDMs have been shown to construct a feature space with structured functional traits and bioclimatic preferences, even though only remote sensing data were provided \citep{deneu_very_2022}.

\subsection{Our indicators originality}
One of the main strengths and originality of our indicators is their scalability.
An analysis can start at the country level with zonal statistics before delving deep into regional patterns.
For example, India ranks fourth in terms of its average proportion of CR species (Table \ref{tab:avg_prop} last column).
Looking at the $\mathcal{I}_{\text{CR}}$ indicator, the Western Ghats and eastern India appear to be the main hosts of CR species.
Finally, the interactive map allows you to zoom in on patterns, explore and look for terrain correspondence with the base maps.
The case study of Sumatra also shows that mountainous regions can host particularly high proportions of CR species.

One of the main shortcomings of our indicators is their lack of transparency. A first direct perspective for improvement is to return, for a given point, the names and IUCN status of the species assemblage. However, this is a technical challenge given the global support size of 221M points.
Another drawback is the interpretability of deep-SDMs. Feature importance experiments would provide a sense of which features the model relies on most. Again, this is a very active area of research and future work will complement this point \citep{ryo_explainable_2021}.

Orchids have specific characteristics that make them valuable indicators of ecosystem health  \citep{newman_orchids_2009}.
They are sensitive to climate change and environmental disturbances \citep{kull2006comparative}, and their interactions with pollinators and mycorrhizal associations contribute to ecosystem functioning \citep{swarts_terrestrial_2009}.
In addition, orchids are easy to monitor in the sense that once a population has been established, it is easy to find it every year.
Therefore, as defined by \citep{jorgensen2016handbook}, orchids can be considered as suitable ecological indicators of ecosystem health.
The family is i) easy to monitor, ii) sensitive to small-scale environmental changes, whose response can be quantified and predicted, and iii) globally dispersed.
They also are umbrella species and their local disappearance may be an early warning of environmental disturbance \citep{gale_orchid_2018}.
However, they don't encompass all aspects of ecosystem biodiversity.
While orchids can be used as surrogate species for biodiversity planning, they can't fully represent overall ecosystem health.
Taking these elements into account, orchid-based indicators such as $\mathcal{I}_\mathcal{O}$ and $\mathcal{I}_c$ can be considered to have a wider scope than just qualifying their family, but also a degree of habitat quality.
Nonetheless, we do not pretend to be able to fully capture ecosystem health through a single family of indicators.
In practice, achieving this goal would require a large number of indicators and measurements.
Comparisons with established indicators are provided in SI \ref{si:Icomp}.

\subsection{Orchids conservation}
Spatial indicators can be used to identify priority areas and support the design of PAs \citep{almpanidou_combining_2021}.
An intuitive method is to select the $k$-highest percentiles of the indicator as hotspots.
In Sumatra, the creation of corridors extending PAs along the Barisan Mountains seems a natural improvement to conserve CR species.
While this approach is easy to understand, there is a risk that some aspects of biodiversity will be missed by the indicator and left unprotected \citep{orme_global_2005}.
It is fair to ask: if the current PAs preserve key aspects of biodiversity and are representative of the other areas identified as most at risk, where is the next priority?
The combination of complementary indicators is the key to designing effective PAs with a limited budget \citep{silvestro_improving_2022}.

Manual extinction risk assessments should be carried out extensively in the tropics and on islands.
Indeed, it is well known that the tropics are poorly assessed, although they host most of the world's biodiversity \citep{collen2008tropical}. The orchid family follows the same trend.
Automated assessment methods will continue to improve, hand in hand with the quality of IUCN assessments in terms of taxonomic coverage, geographical extent and consistency.
Finally, special attention must be paid to the assessment and protection of islands: all our indicators point to them as hosts of particularly threatened species assemblages.

\subsection{Conclusions}
Based on deep-SDMs architectures, we have developed global indicators that qualify the extinction risk of species assemblages at an unprecedented kilometre resolution.
This allows multiscale analysis from global patterns down to country statistics or landscape discrepancies.
The indicators are available as interactive maps at \url{https://mapviewer.plantnet.org/?config=apps/store/orchid-status.xml#}.
Although our results show how our novel indicators can be successfully employed, working closely with decision-makers would ultimately allow for more effective guidance of conservation actions \citep{guisan_predicting_2013}.
To enable efficient technology transfer, interdisciplinary studies between computer science and conservation science need dialogue with conservation practitioners \citep{gale_orchid_2018}.

\section*{Acknowledgements}
The research described in this paper was funded by the European Commission via the GUARDEN and MAMBO projects, which have received funding from the European Union’s Horizon Europe research and innovation programme under grant agreements 101060693 and 101060639.
The opinions expressed in this work are those of the authors and are not necessarily those of the GUARDEN or MAMBO partners or the European Commission.
The INRIA exploratory action CACTUS fund also supported this work.
This work was granted access to the HPC resources of IDRIS under the allocation 20XX-AD011013648 made by GENCI.
Finally, we warmly thank Alexander Zizka for providing us with the filtered set of orchid occurrences.

\bibliographystyle{apalike}
\bibliography{sample}
\clearpage

\appendix
\section*{Supplementary information}

\renewcommand{\thefigure}{S\arabic{figure}}
\renewcommand{\thetable}{S\arabic{table}}
\renewcommand{\theequation}{s\arabic{equation}}
\setcounter{figure}{0}
\setcounter{table}{0}
\setcounter{equation}{0}

\section{Deep neural network architecture, dataset spatial split and training procedure}
\label{si:archi_training}

\paragraph{Inception v3}
Our backbone model is an adaptation of the Inception v3 \citep{szegedy_rethinking_2016}. Initially designed to accept three-channel rgb images, it was modified to deal with a higher number of channels.
This convolutional neural network learn patterns from spatialized input predictors. Letting models benefit from the spatial information was shown successful in various literature applications \citep{botella2018deep, deneu_convolutional_2021}.
Successive inception modules are composed of convolutional filters of different sizes. This allows the different patch patterns of all sizes to be captured.
Convolutional layers reduce the very high input dimension and a final softmax layer outputs the conditional probability distributions.
Inputs are concatenated along the channel dimension. It results in $N \times 64 \times 64$ tensors with $N$ the total channel number. Pixel resolution is of 1~km. A large $64 \times 64$~km² environmental context is therefore provided.
The model is also spatially explicit: observation longitude and latitude are supplied in two dedicated channels along with the other predictors.\\

Deep learning models successfully process large numbers of inputs and classes with few samples. In fact, the modelling paradigm is completely different from combined per-species models.
The filters learned during training are applied to all samples, all classes combined.
The final softmax layer, which outputs class probabilities conditionally on an observation, is based on a reduced representation space common to all classes.
This space has been shown to be structured by the ecological preferences of species in \citep{deneu_very_2022}.
More generally, deep learning classification with strong class imbalance is a very active research avenue.
DL outperforms classical approaches to model classes with few samples.
In conclusion, our deep-SDM is not affected by the curse of dimensionality.\\

\paragraph{Dataset spatial split}
A spatial block hold-out validation strategy is employed to limit the effect of the spatial auto-correlation in the data in the evaluation of the model (as suggested in \cite{roberts2017cross}). 0.025° longitude / latitude blocks were defined worldwide (equivalent to 2.8~km at the equator).
A train/validation/test split of 90/5/5~\% of the blocks is then applied.
The split is further stratified to \textit{WGSRPD} level 2 zones to ensure a balanced block distribution across vegetation units - \textit{World Geographical Scheme for Recording Plant Distributions}, \cite{brummitt2001world}.
In order not to over penalise performance, species initially present only in the validation or test sets are transferred to the training set.
At the occurrence level, this results in a 902,174 / 46,290 / 50,794 set distribution.
At the species level, this leads to a 14,129 / 4,037 / 4,166 split.

\paragraph{Training procedure}
The deep neural network $f_{\theta}$ is trained with the widely recognised LDAM loss, a modified cross-entropy function giving more emphasis to rare species during the training \citep{cao_learning_2019}. It is a \textit{label-distribution-aware} function specifically designed for strong class-imbalance and multi-class classification problems. Performances on rare species are pushed upward without deteriorating common species predictions.

Model is fitted on Jean Zay, a supercomputer from the Institute for the Development and Resources in Intensive Scientific Computing (IDRIS).
Layer weights are initialized from a truncated normal continuous random variable.
Stochastic gradient descent optimizes the parameters on 2 GPUs during 70 epochs. With a batchsize equal to 128 and an initial learning rate of 0.01, the training process took 45~h. Learning rate is decayed at epochs 50 and 65 by a ten factor. A trained model weighs 610~MB.
After having validated the model, a new training is lead from scratch on the whole dataset. It is stopped at the best epoch determined beforehand on the validation set. This retraining aims at obtaining the best possible model weights before the global-scale inference.
\clearpage

\section{Species assemblage post-processing}
\label{si:oor}
\begin{figure}[H]
    \vspace{-0.25cm}
    \begin{center}
    \frame{\includegraphics[width=\textwidth]{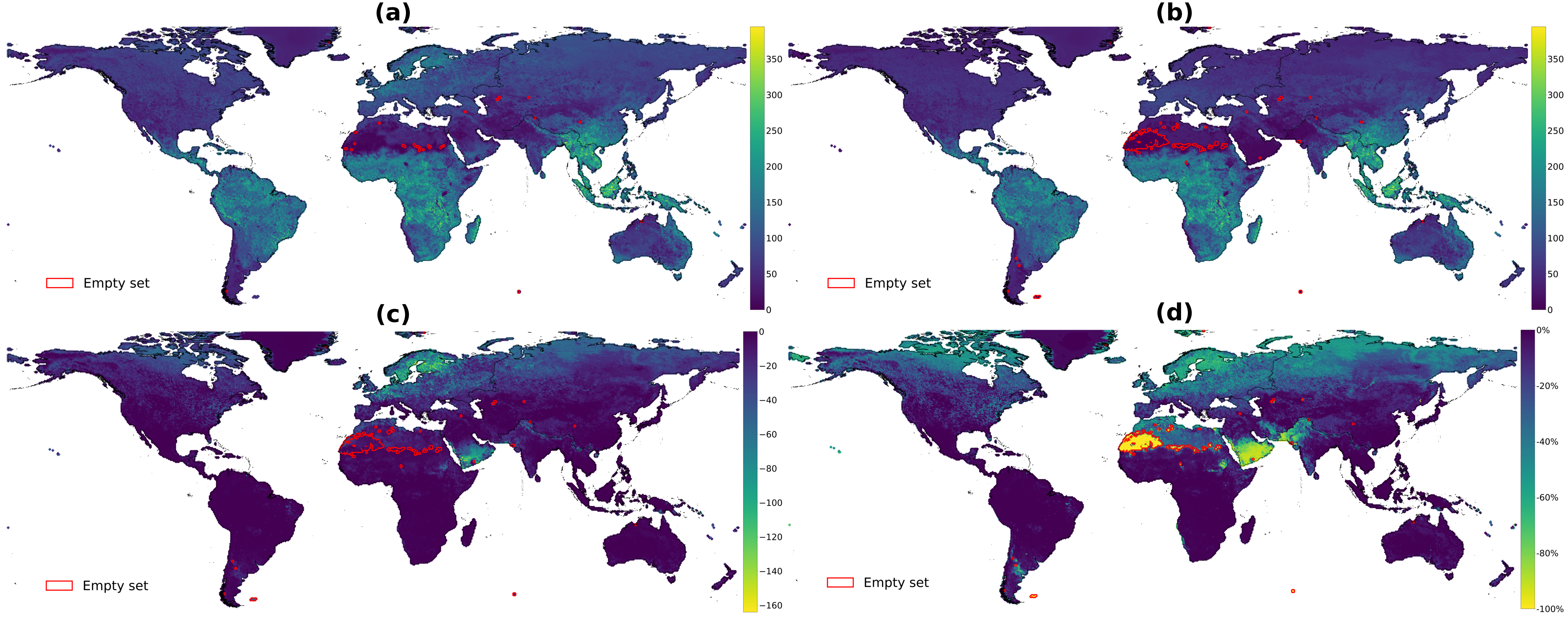}}
    \vspace{-0.25cm}
    {\small \textbf{(e)}}
    \begin{threeparttable}

\rowcolors{2}{gray!25}{white}
\begin{tabular}{lrr}
\headrow
{} & $\Delta$ \textbf{species} & \textbf{Relative change [\%]} \\
\textbf{mean} & -14.35 & -18.21 \\
\textbf{std} & 19.37 & 21.28 \\
\textbf{min} & -164 & -100.0 \\
\textbf{25\%} & -22 & -31.3 \\
\textbf{50\%} & -6 & -9.1 \\
\textbf{75\%} & -1 & -0.7 \\
\textbf{max} & 0 & 0.0 \\
\hline
\end{tabular}
\end{threeparttable}

    \end{center}
    \vspace{-0.25cm}
    \caption{
    Maps and statistics illustrating the post-filtering step with the geographic-prior.
    The support is a global regular grid with $0.5$ decimal degree resolution (59,823 points).\\
    (a) $|\hat{S}_\lambda|$, i.e. the species assemblage size \textit{before} the filtering step.
    Northern latitudes -and especially northern Europe- present abnormally large species assemblages.
    This is a consequence of the generalisation / over-prediction trade-off described in Discussion.
    The prediction model is over-confident because of the extensive occurrence training data in northern European countries.\\
    (b) $|\hat{S}'_\lambda|$, i.e. the species assemblage size \textit{after} the filtering step.
    The over-prediction bias at northern latitudes has been largely compensated.
    Empty predictions zones (red surrounded) have increased because of the geographic filtering, especially in the Sahara.\\
    (c) $|\hat{S}'_\lambda| - |\hat{S}_\lambda|$, i.e. the absolute size difference of the species assemblage \textit{before/after} the filtering step.
    Regions having lost the highest number of species are northern European countries and  the South Arabian Peninsula.\\
    (d) $\frac{|\hat{S}'_\lambda| - |\hat{S}_\lambda|}{|\hat{S}_\lambda|}$, i.e. the relative change in the species assemblage size \textit{before/after} the filtering step.
    Regions mentioned in (c) are highlighted again.
    Saharan regions with empty predictions after geo-filtering do not appear to have lost high species number in (c).
    However, the clear yellow on map (d) indicates that these regions have lost all of the few species they were predicted to host.\\
    (e) Statistics on the absolute and relative size difference of the species assemblage \textit{before/after} geo-filtering.
    $\Delta$ \textbf{species} corresponds to map (c) and \textbf{Relative change [\%]} corresponds to map (d).
    }\label{si:oor_fig}
\end{figure}

Species assemblages and their relative probabilities are finally post-processed in two steps.
We derive first from the initial occurrence set the inhabited continents of each species (WGSRPD level 1). This allows to filter out species predicted by the model outside their known continents of presence.
Filtered assemblages of species are denoted $\hat{S}'_\lambda$.
We computed statistics on this filtering step with a geographic prior on a global regular grid with $0.5$ decimal degree resolution. The median number of species removed is $6$, or $9.1$\% of the assemblage. Full statistics and map discrepancies are further discussed in Figure \ref{si:oor_fig}.
Second, kept species conditional probabilities are normalised. Species with a conditional probability of presence smaller than $\lambda$ are considered absent from the predicted assemblage, as well as species predicted outside their known continents of presence. In both cases, associated conditional probabilities were forced to zero. Normalisation allows to get back to a probability distribution summing to one.
For a given input $x$, final probabilities are obtained with $\hat{\eta}'(x) = \frac{\hat{\eta}(x)}{\sum\limits_{l \in \hat{S}'_\lambda(x)} \hat{\eta}_{l}(x)}$.
\clearpage

\section{Model evaluation and calibration}
\label{si:modelEval}
\paragraph{Evaluation of the deep-SDM}
The deep-SDM model was evaluated on unseen occurrences from the validation spatial blocks.
Validation performances set the best epoch choice - the 69$^{th}$ - for final test set metrics to be computed.
Selected metrics are the \textit{top-$k$ accuracy} and its per-class counterpart the \textit{top-$k$ accuracy per species}.
These set-valued metrics do not require pseudo absences to avoid potential induced bias \citep{phillips2009sample}.
Top-$k$ accuracy measures if the model returns the correct label among the $k$ most likely classes:
\begin{equation}
    \label{equ:topk}
    \text{A}_{k}(i) = \begin{cases}
      1 & \text{if $\hat{\eta}_{y_i}(x_i)\geq \Tilde{\eta}_{k}(x_i) $}\\
      0 & \text{otherwise}
    \end{cases}
\end{equation}
with $(x_i,y_i)$ an input/label pair and $\Tilde{\eta}$ the permutation of $\hat{\eta}$ sorted in descending order.\\
The success rate can be calculated for all test set occurrences, all classes combined (\textit{micro-average} denoted $\text{A}_{k}$) or first for each class individually and then averaged together (\textit{macro-average} denoted $\text{MSA}_{k}$). The former gives prominence to common species by construction, while the latter depends heavily on rare species performances. Macro-average metrics are suitable for highly imbalanced datasets.

\begin{figure}[H]
    \begin{center}
    \frame{\includegraphics[width=\textwidth]{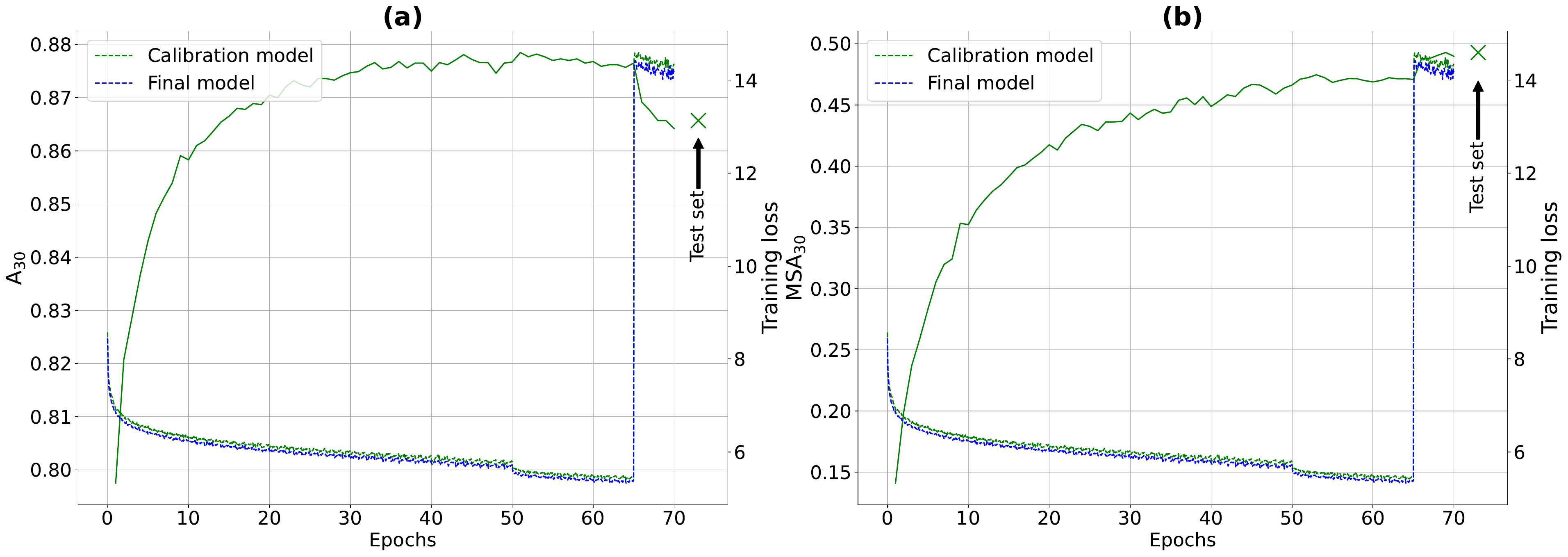}}
    \end{center}
    \caption{
    \textit{Solid lines}: (a) micro-average and (b) macro-average top-30 accuracy on the validation set over training.
    \textit{Dotted lines}: training losses of the calibration model (only on the training set) and the final model (full dataset).
    The first loss jump at epoch 50 corresponds to the first learning rate decay by a tenth factor.
    The second loss jump at epoch 65 results from the LDAM delayed reweighting scheme.
    It gives prominence to the rare species, which is why we observe a decrease in performance for $\text{A}_{30}$ and an increase for $\text{MSA}_{30}$.
    The training and validation curves show no signs of overfitting.
    The performances of the test sets with the best calibration model (epoch 69) are $\text{A}_{30} = 0.865$ and $\text{MSA}_{30} = 0.48$.
    }\label{si:loss_fig}
\end{figure}

\begin{figure}[H]
    \begin{center}
        \frame{\includegraphics[width=13cm]{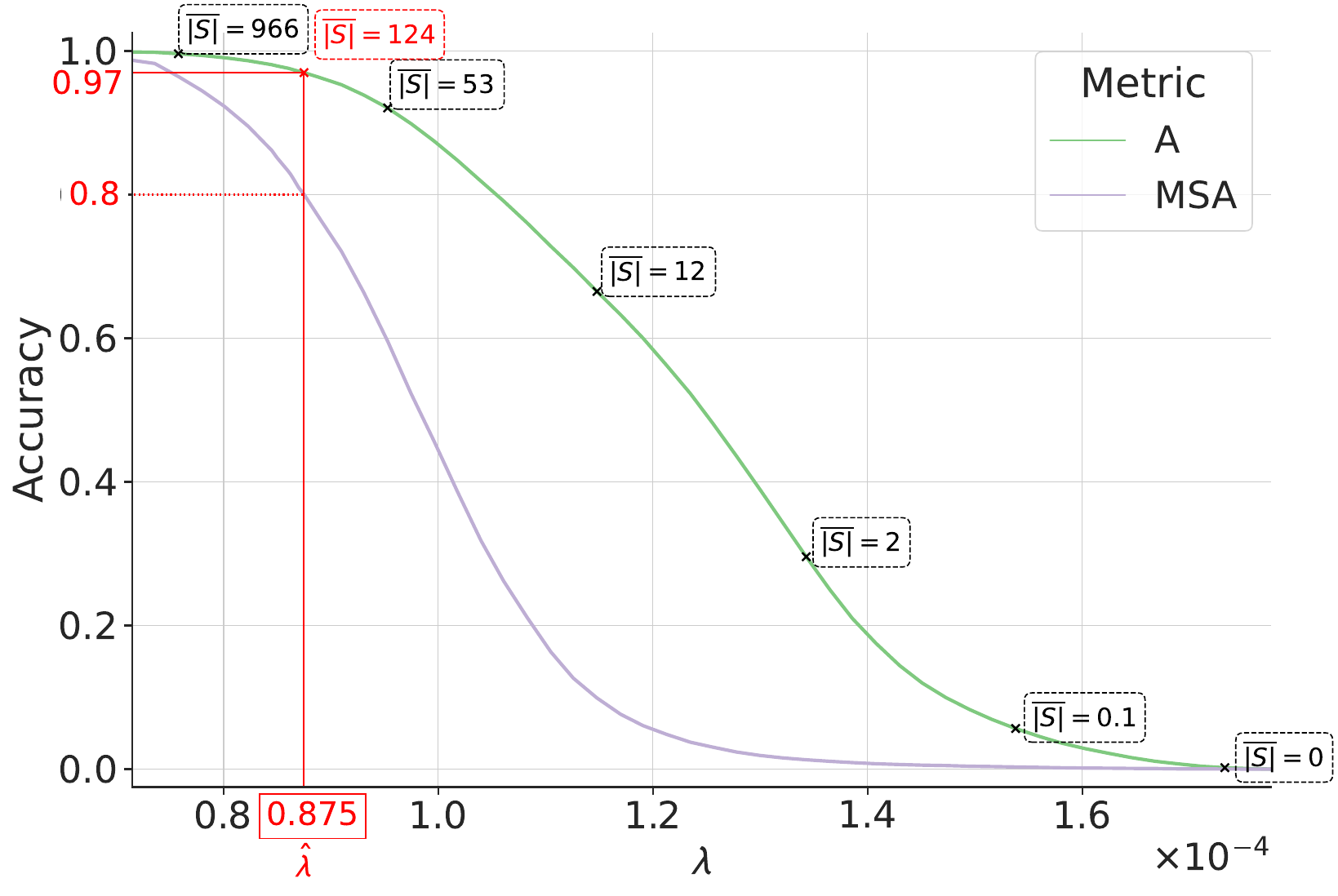}}
        \caption{ 
        Average error control setting on the validation set. Limiting condition on the micro-average accuracy (green curve) is  $\epsilon \leq 0.03 \Leftrightarrow \text{A} \geq 0.97$.  Optimal threshold $\hat{\lambda}$ is highlighted in red while matching macro-average accuracy (grey function) is also reported with a red dashed line. Average set sizes $\overline{|\hat{S}_\lambda|}$ are indicated in dashed boxes (hat and subscript being dropped for readability).
        }\label{si:model_calib}
    \end{center}
\end{figure}

Final test set performances at epoch 69 are $\text{A}_{30} = 0.87$ and $\text{MSA}_{30} = 0.48$.
This means that i) the correct label is returned among the first 30 species for 87\% of the test observations (representative of common species), and ii) when each species in the test set is given the same weight, the correct label is within the first 30 classes returned almost half the time.
This second metric may seem low, but it actually measures a particularly difficult task, given that the test set contains 4,166 species $(30/4166 \leq 1\%)$.
Furthermore, it reflects the performance of the model on rare species, and Figure \ref{si:model_calib} shows that considering \textit{on average} 124 species significantly improves performance on the validation set, see next paragraph.
Finally, training and validation curves show no sign of overfitting, see Figure \ref{si:loss_fig}.\\

\begin{figure}[!h]
    \vspace{-0.25cm}
    \hspace{3.5cm} \textbf{(a)} \hspace{6.cm} \textbf{(b)}
    \vspace{-0.3cm}
    \begin{center}
    \includegraphics[trim={0.2cm 0 0.3cm 0}, clip, width=6.5cm]{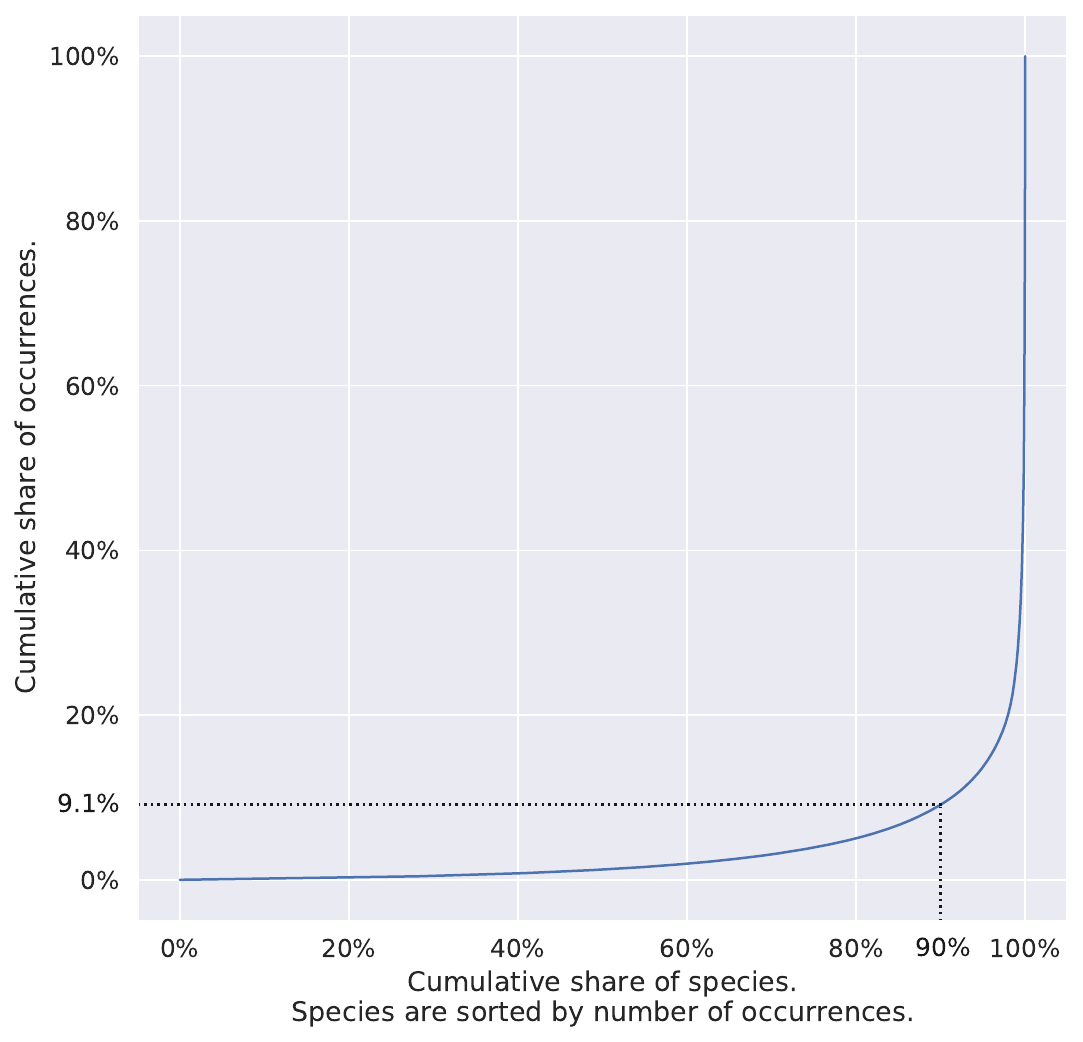}
    \includegraphics[trim={0.2cm 0 0.3cm 0}, clip, width=6.5cm]{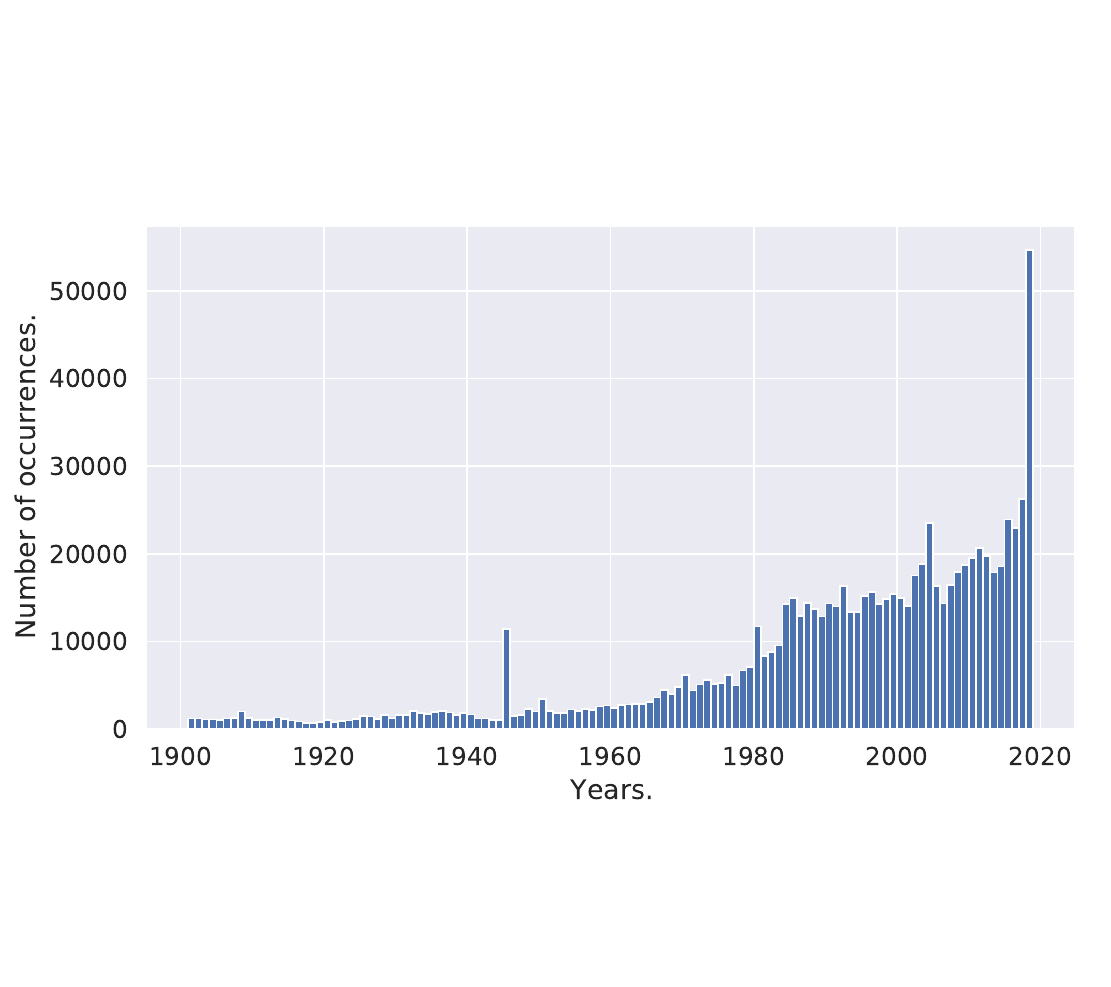}
    \end{center}
    
    \hspace{0.2cm}\textbf{(c)}
    \vspace{-1cm}
    \begin{center}
    \includegraphics[width=10cm]{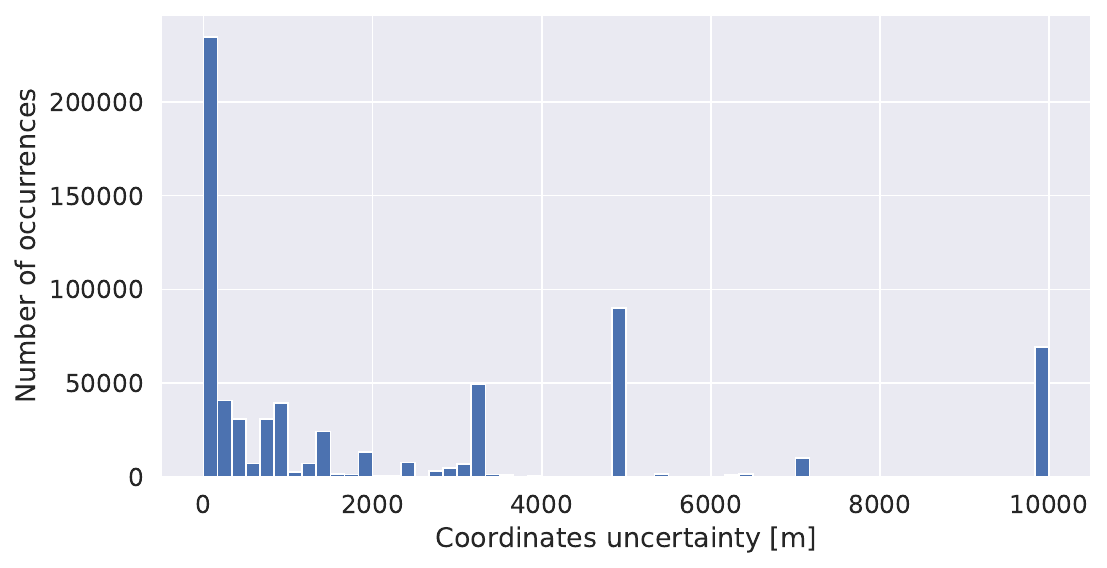}
    \end{center}
    
    \hspace{0.2cm}\textbf{(d)}
    \vspace{-1cm}
    \begin{center}
    \includegraphics[trim={3cm 2.5cm 13cm 1.8cm}, clip, width=12.25cm]{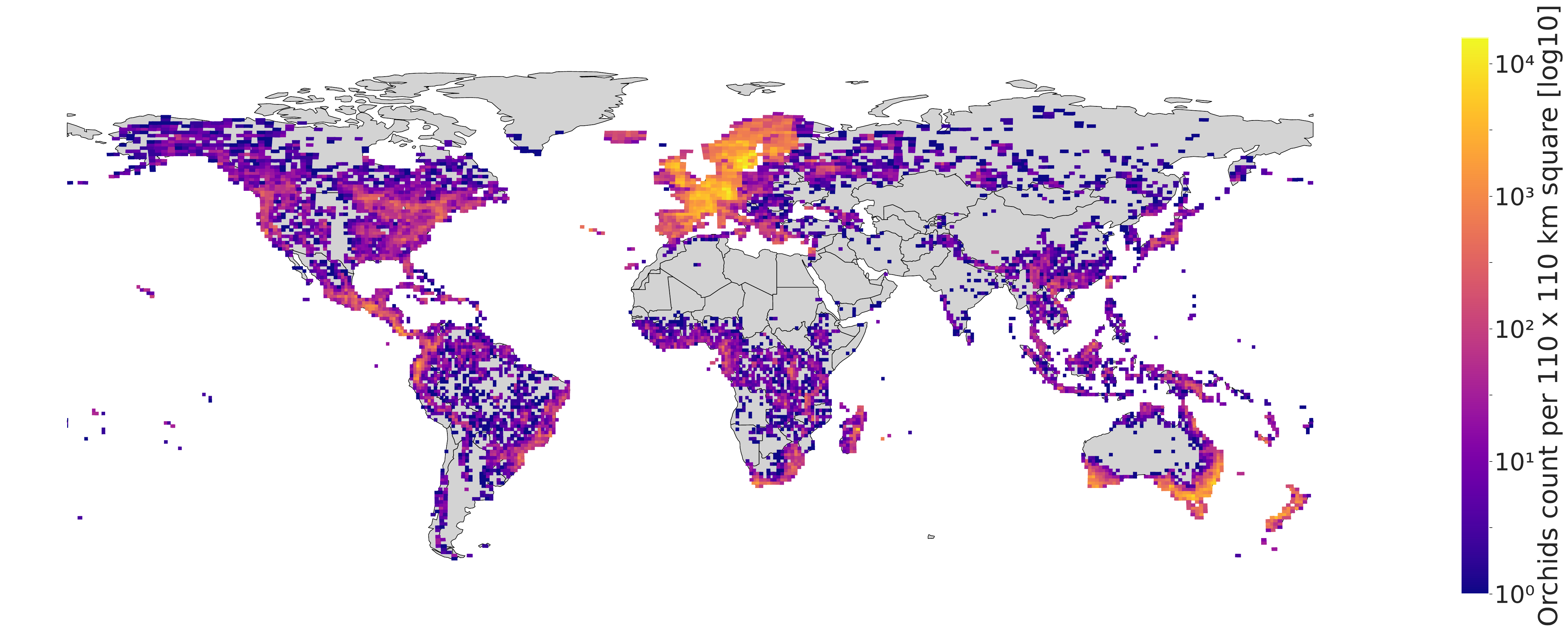}
    \includegraphics[trim={76cm 0 0 0}, clip, width=0.8cm]{Orchids_map_newleg.pdf}
    \end{center}
    
    \vspace{-0.5cm}
    \caption{
    \textbf{(a)} Occurrences' distribution. Species are ordered by frequency. The dotted lines are flagging that 90\% of the species are only gathering 9.1\% of the occurrences.
    \textbf{(b)} Occurrences' temporal distribution. The two previous graphs are based on all dataset's occurrences.
    \textbf{(c)} Histogram of occurrences geolocation uncertainty (60 bins). 31\% of the 999,248 occurrences associated with satellite data had no uncertainty provided at all and are not represented in this figure. Uncertainty was limited to 10,000~m on the Figure. First quartile is 100~m, median is 850~m and third quartile is 5,000~m. Recent and citizen science occurrences are usually integrating quite precise geolocation (explaining left peaks accumulation) whereas old observations will be less precise. The peak at 5,000~m certainly witnesses an arbitrary uncertainty value attributed to part of the orchids.
    While georeferencing uncertainty is effectively a method limitation, deep learning algorithms can statistically learn correct class representations even in the presence of noise \citep{elith_novel_2006, rolnick_deep_2018}.
    \textbf{(d)} Observations map coloured by number of records in 110 x 110~km tiles (log10 scale).
    }\label{si:distrs_fig}
\end{figure}

\paragraph{Calibration of the species assemblage prediction model}
As discussed earlier, the optimisation of the hyper-parameter $\lambda$ is done through an average error control method applied on the validation set. In Equation 3, $\epsilon$ is set to $0.03$.
The resulting estimated value for $\lambda$ is equal to $8.75\mathrm{e}{-5}$ (see Figure \ref{si:model_calib}) and the corresponding average size of the predicted species assemblages is equal to $124$ species.
Reaching 0.97\% micro-average accuracy means that the model almost always returns the correct label within the predicted set when a random unseen observation is being provided. The number of observations per class being strongly unbalanced (see Fig. \ref{si:distrs_fig}a, the 97\% micro-average accuracy is strongly influenced by the performance on common species.
Now, when all unseen species are granted the same weight in the average computation (macro-average accuracy), performance is still of 80\%. Given how unbalanced the observation dataset is (median occurrence number is four, 25\% species have more than 13 occurrences), it becomes clearer that the model's performances are satisfying. 
Summary statistics on $|\hat{S}_\lambda|$ are reported on Table \ref{si:Tstats}.\\

\begin{table}
    \begin{center}
        \caption{ 
        Validation set statistics on $|\hat{S}_\lambda|$, i.e. the size of the species assemblage after thresholding the conditional probabilities of presence with $\hat{\lambda}$ (46,290 validation points).
        The minimum number of species retained in the validation set is four.
        However, on a global scale there are areas with no species above $\hat{\lambda}$, resulting in empty predictions (e.g. western Algeria).
        It is also very likely that areas are predicted with more than 401 orchids (the maximum on the validation set).
        }\label{si:Tstats}
        \begin{threeparttable}
            \rowcolors{2}{gray!25}{white}
            \begin{tabular}{lrrrrrrrrr}
                \headrow
                {} & \textbf{mean} & \textbf{std} & \textbf{min} & \textbf{25\%} & \textbf{50\%} & \textbf{75\%} & \textbf{max} & \textbf{A}$_{30}$ & \textbf{MSA}$_{30}$\\
                \textbf{$\hat{\lambda}$} &  124.067 &  39.803 &    4 &   95 &  121 &  150 &  401 &               0.970 &                0.801 \\
                \hline
            \end{tabular}
        \end{threeparttable}
    \end{center}
\end{table}
\clearpage

\section{Maps batch processing and online access}
\label{si:maps_batch_and_access}
\paragraph{Batch processing}
Species assemblages are predicted by 50,176 batches for volume reasons. PyTorch model 512-size predictions are accumulated in a buffer until exceeding the 50,000 limit and are then exported. Raw predicted classes and probabilities are appended in binary files whereas spatial indicators are computed on the fly and saved in distinct .geotiff format. Finally, geotiffs are merged and converted to the Cloud Optimized GeoTIFF format (COG): 256 x 256 blocks are tiling the data and six levels of overviews are added.\\

\paragraph{Online and interactive map access}
The web mapping solution used to render the raster data in a web environment is quite simple.
It relies upon a serverless front-end application and a map server able to render OGC (Open Geospatial Consortium) compliant web services : Web Map Service, Web Feature Service, etc.
The front-end application is an open-source project called MViewer (\url{https://mviewer.netlify.app/fr/}) and is mostly implemented by Brittany region. It parses an .xml configuration file to generate an interactive map. The map server is also an open-source project : GeoServer (\url{https://geoserver.org/}). It can read COG data among other geospatial data and serves it as a web service to be displayed by the front-end application.
Website is available at \url{https://mapviewer.plantnet.org/?config=apps/store/orchid-status.xml#}.
\clearpage

\section{Orchid dataset distributions}
\label{si:occ_distributions}

\begin{figure}
    \begin{center}
        \includegraphics[trim={3cm 1cm 3cm 1cm}, clip, width=\textwidth]{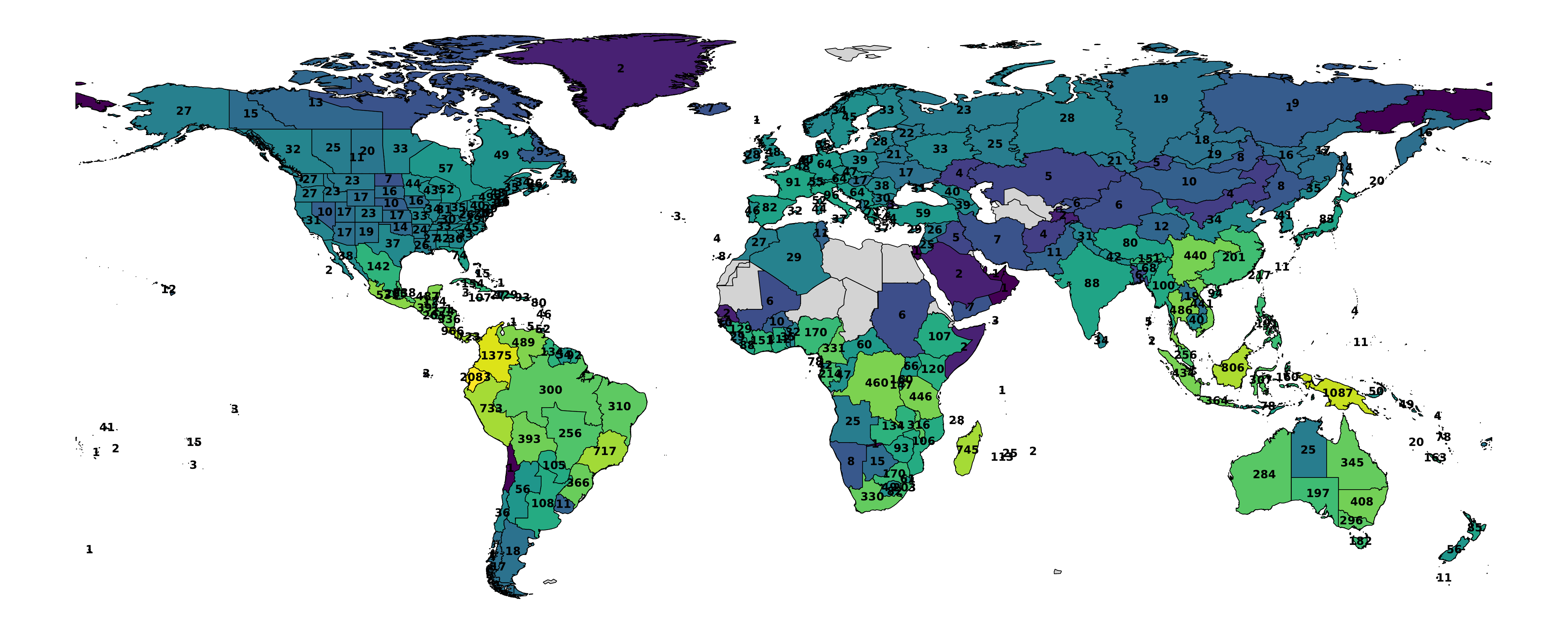}
        \caption{Species richness map stratified by botanical country (WGSRPD level 3). Colours are in log scale.
        }\label{si:richness_map}
    \end{center}
\end{figure}

\begin{figure}
\hspace{1cm} \textbf{(a)} \hspace{6.5cm} \textbf{(b)}
\vspace{-0.3cm}
\centering
\begin{center}
    \includegraphics[trim={0cm 1.45cm 0cm 0cm}, clip, width=0.48\textwidth]{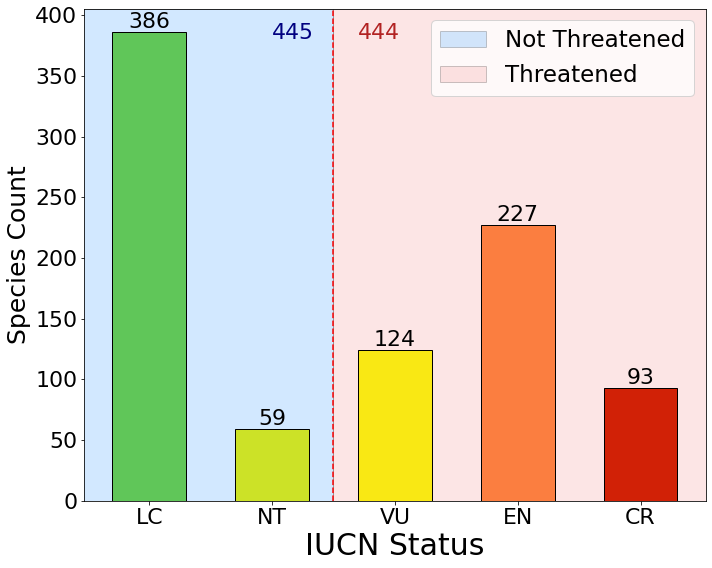}
    \includegraphics[trim={0cm 1.45cm 0cm 0cm}, clip, width=0.48\textwidth]{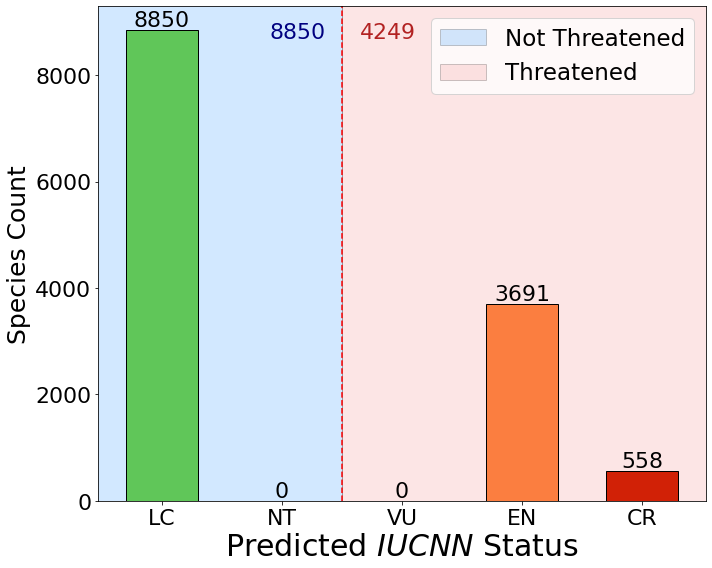}
\end{center}
\caption{
Distribution of extinction risk status of (a) known IUCN-assessed species and (b) predicted with \textit{IUCNN} (Zizka et al., 2021).
No NT or VU species are predicted. These two classes are by definition quite ambiguous: a species is classified as Near Threatened if the VU thresholds are almost met. When optimising the classifier, predicting NT or VU species results in a high error rate, so the best performance trade-off in this setting is to exclude these classes.
}\label{si:status_distrs}
\end{figure}
\clearpage

 \section{Predictive features description}
\label{si:predictors}
In selecting predictive features, the main limiting criterion was to use only globally available potential drivers of orchid preferences.

\paragraph{WorldClim2 bioclimatic variables}
The nineteen standard bioclimatic variables from WorldClim version 2 were provided to the model \citep{fick_worldclim_2017}. They are historic averages over the 1970-2000 at 30-second resolution. This suits our occurrence date distribution. Variables stem from temperature and precipitation data (\url{https://www.worldclim.org/data/bioclim.html}). They are established indicators of climate annual trends, seasonality and extreme values.

\paragraph{Soilgrids pedological variables}
Soilgrids is a collection of eleven global soil property and class maps produced by machine learning models \citep{poggio2021soilgrids}. They include soil pH, nitrogen concentration annd clay particles proportions among others (more information in the official FAQ: \url{https://www.isric.org/explore/soilgrids/faq-soilgrids}). The exploited statistical models are fitted with 230,000 soil profiles spread wordwide and environmental covariates. We use the 1-kilometre resolution products.

\paragraph{Human footprint detailed rasters}
Eight variables measure direct and indirect global human pressure: built environments, population density, electric infrastructure, crop lands, pasture lands, roads, railways, and navigable waterways \citep{venter_global_2016}. They are provided at a 1-kilometre resolution (\url{https://datadryad.org/stash/dataset/doi:10.5061/dryad.052q5}) and for two distinct years: 1993 and 2009. These rasters spring form both remotely-sensed data and surveys.

\paragraph{Terrestrial ecoregions of the world}
This is a biogeographic classification of terrestrial biodiversity. Ecoregions are defined by the authors as \textit{"relatively large units of land containing a distinct assemblage of natural communities and species, with boundaries that approximate the original extent of natural communities prior to major land-use change"} \citep{olson_terrestrial_2001}. 867 ecoregions are gathered into 14 biomes such as boreal forests or deserts. Data (\url{https://www.worldwildlife.org/publications/terrestrial-ecoregions-of-the-world}) was resampled at 30 seconds longitude/latitude resolution.

\paragraph{Location}
The explicit provision of observation coordinates is a key modelling decision. Both a large regional context and precise location information are provided. The model can make the most of this mixed input. Deep learning models can indeed take advantage of complex combinations of heterogeneous inputs. As longitude and latitude are inputted separately, both indications are processed alongside, can interact, but are also interpreted distinctly.

\begin{figure*}[!htb]
\textbf{(a)}\\
{\centering
\includegraphics[width=\textwidth]{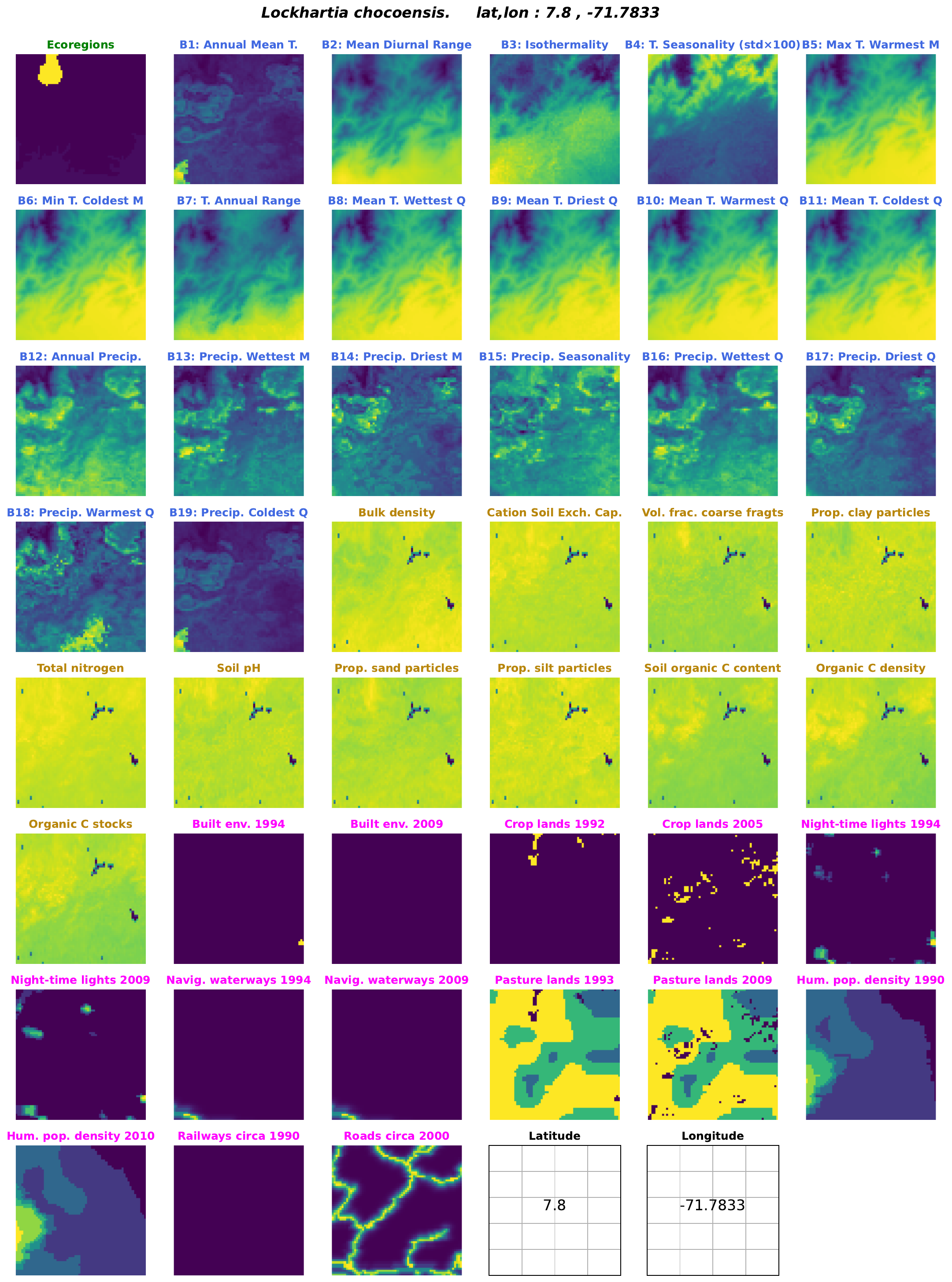}}
\captionsetup{labelformat=empty}
\end{figure*}

\begin{figure*}[!htb]
\textbf{(b)}\\
{\centering
\includegraphics[width=\textwidth]{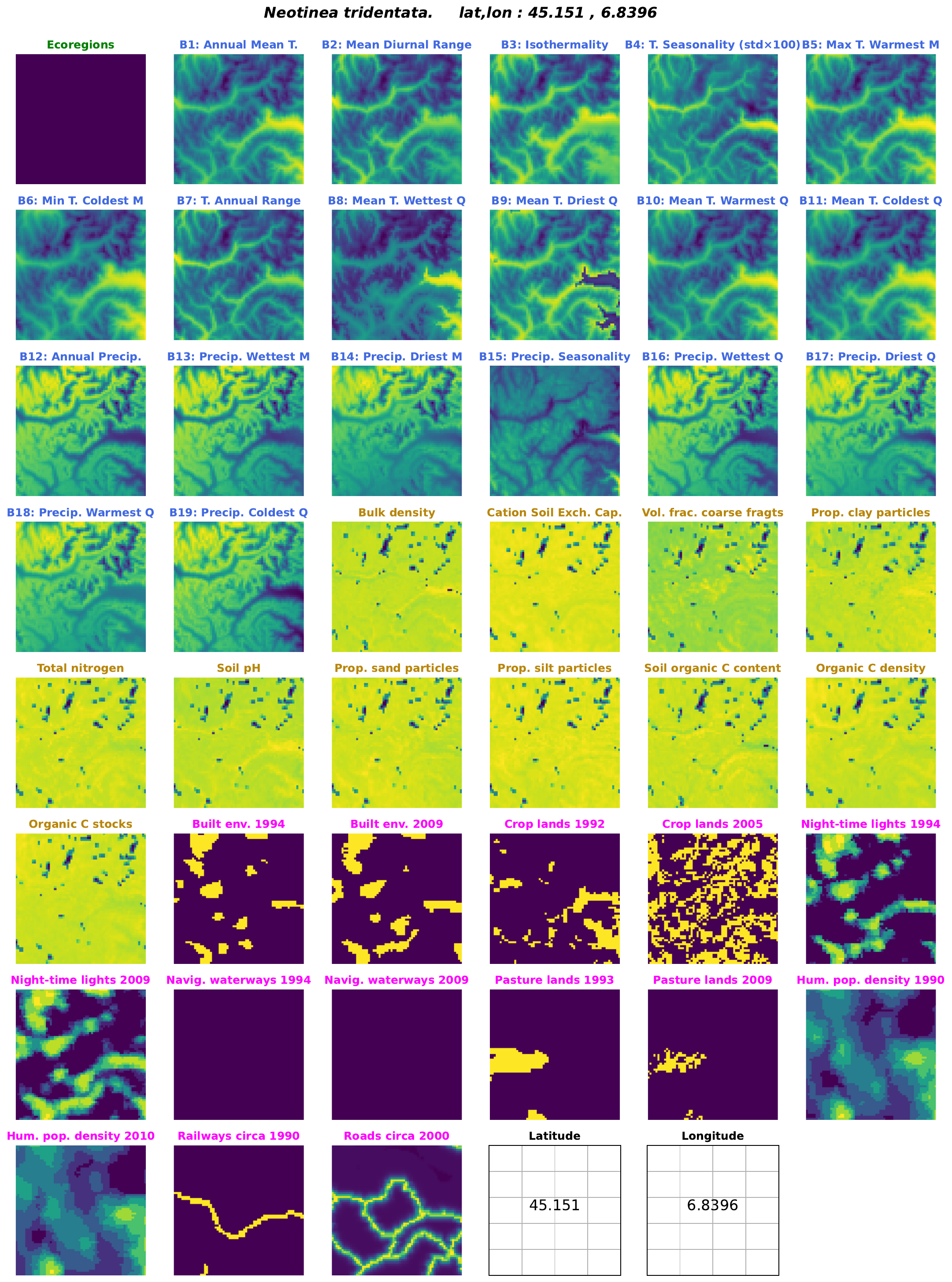}}
\captionsetup{labelformat=empty}
\end{figure*}

\setcounter{figure}{6}
\begin{figure}[!htb]
\textbf{(c)}\\
{\centering
\includegraphics[width=\textwidth]{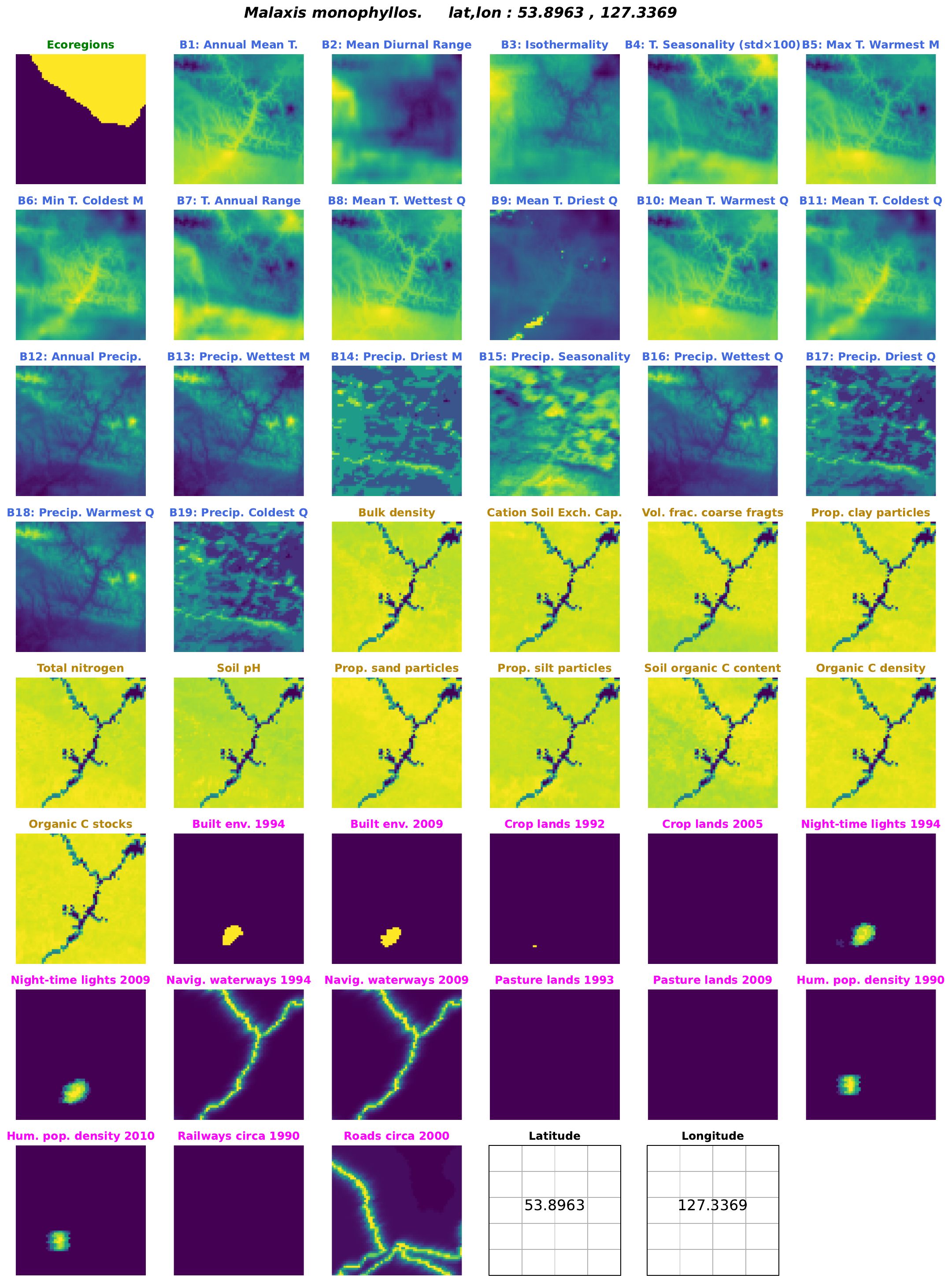}}
\caption{
2D input data associated to three observations located in (a) Venezuela, (b) France and (c) Russia.
Feature types are denoted by their title color:
\textit{(green)} ecoregions, \textit{(blue)} bioclimatic variables, \textit{(brown)} pedological variables, \textit{(pink)} human footprint and \textit{(black)} location.
}\label{si:input_exs}
\end{figure}
\clearpage

\begin{table}[H]
\centering
\scriptsize
\begin{threeparttable}
\caption{
List of predictors. They are either categorical or continuous and can be gathered into five groups: the terrestrial ecoregions of the world, the WorldClim2 bioclimatic variables, the Soilgrids pedological variables, the detailed rasters of the human footprint and the location.
}\label{tab:features_table}

\rowcolors{2}{gray!25}{white}
\begin{tabular}{llll}
\headrow
& \textbf{Group} & \textbf{Name} & \textbf{Type}\\
\textbf{1 } &  Terrestrial ecoregions of the world &                               Ecoregions per biome &  categorical \\
\textbf{2 } &     WorldClim2 bioclimatic variables &                     BIO1 = Annual Mean Temperature &   continuous \\
\textbf{3 } &       &  BIO2 = Mean Diurnal Range (Mean of monthly (max temp - min temp)) &   continuous \\
\textbf{4 } &       &            BIO3 = Isothermality (BIO2/BIO7) (×100) &   continuous \\
\textbf{5 } &       &  BIO4 = Temperature Seasonality (standard deviation ×100) &   continuous \\
\textbf{6 } &       &            BIO5 = Max Temperature of Warmest Month &   continuous \\
\textbf{7 } &       &            BIO6 = Min Temperature of Coldest Month &   continuous \\
\textbf{8 } &       &        BIO7 = Temperature Annual Range (BIO5-BIO6) &   continuous \\
\textbf{9 } &       &         BIO8 = Mean Temperature of Wettest Quarter &   continuous \\
\textbf{10} &       &          BIO9 = Mean Temperature of Driest Quarter &   continuous \\
\textbf{11} &       &        BIO10 = Mean Temperature of Warmest Quarter &   continuous \\
\textbf{12} &       &        BIO11 = Mean Temperature of Coldest Quarter &   continuous \\
\textbf{13} &       &                       BIO12 = Annual Precipitation &   continuous \\
\textbf{14} &       &             BIO13 = Precipitation of Wettest Month &   continuous \\
\textbf{15} &       &              BIO14 = Precipitation of Driest Month &   continuous \\
\textbf{16} &       &  BIO15 = Precipitation Seasonality (Coefficient of Variation) &   continuous \\
\textbf{17} &       &           BIO16 = Precipitation of Wettest Quarter &   continuous \\
\textbf{18} &       &            BIO17 = Precipitation of Driest Quarter &   continuous \\
\textbf{19} &       &           BIO18 = Precipitation of Warmest Quarter &   continuous \\
\textbf{20} &       &           BIO19 = Precipitation of Coldest Quarter &   continuous \\
\textbf{21} &      Soilgrids pedological variables &                              Bulk density (cg/cm3) &   continuous \\
\textbf{22} &        &      Cation exchange capacity at ph 7 (mmol(c)/kg) &   continuous \\
\textbf{23} &        &                        Coarse fragments in cm3/dm3 &   continuous \\
\textbf{24} &        &                               Clay content in g/kg &   continuous \\
\textbf{25} &        &                                  Nitrogen in cg/kg &   continuous \\
\textbf{26} &        &                                  pH water (pH *10) &   continuous \\
\textbf{27} &        &                                       Sand in g/kg &   continuous \\
\textbf{28} &        &                                       Silt in g/kg &   continuous \\
\textbf{29} &        &                        Soil organic carbon (dg/kg) &   continuous \\
\textbf{30} &        &                     Organic carbon density (g/dm3) &   continuous \\
\textbf{31} &        &                   Soil organic carbon stock (t/ha) &   continuous \\
\textbf{32} &     Human footprint detailed rasters &  Individual pressure map of built environments in 1994 &  categorical \\
\textbf{33} &       &  Individual pressure map of built environments in 2009 &  categorical \\
\textbf{34} &       &      Individual pressure map of crop lands in 1992 &  categorical \\
\textbf{35} &       &      Individual pressure map of crop lands in 2005 &  categorical \\
\textbf{36} &       &  Individual pressure map of night-time lights in 1994 &  categorical \\
\textbf{37} &       &  Individual pressure map of night-time lights in 2009 &  categorical \\
\textbf{38} &       &  Individual pressure map of navigable waterways in 1994 &   continuous \\
\textbf{39} &       &  Individual pressure map of navigable waterways in 2009 &   continuous \\
\textbf{40} &       &   Individual pressure map of pasture lands in 1993 &  categorical \\
\textbf{41} &       &   Individual pressure map of pasture lands in 2009 &  categorical \\
\textbf{42} &       &  Individual pressure map of human population density in 1990 &  categorical \\
\textbf{43} &       &  Individual pressure map of human population density in 2010 &  categorical \\
\textbf{44} &       &     Individual pressure map of railways circa 1990 &  categorical \\
\textbf{45} &       &        Individual pressure map of roads circa 2000 &   continuous \\
\textbf{46} &                             Location &                                     Longitude (DD) &   continuous \\
\textbf{47} &                              &                                      Latitude (DD) &   continuous \\
\hline
\end{tabular}

\end{threeparttable}
\end{table}
\clearpage

\section{Most threatened and diverse orchid assemblages}
\label{si:venn_table}
\begin{figure}[!htb]
{\centering
\textbf{(a)}
\includegraphics[trim={0cm 0cm 0cm 0.75cm}, clip, width=\textwidth]{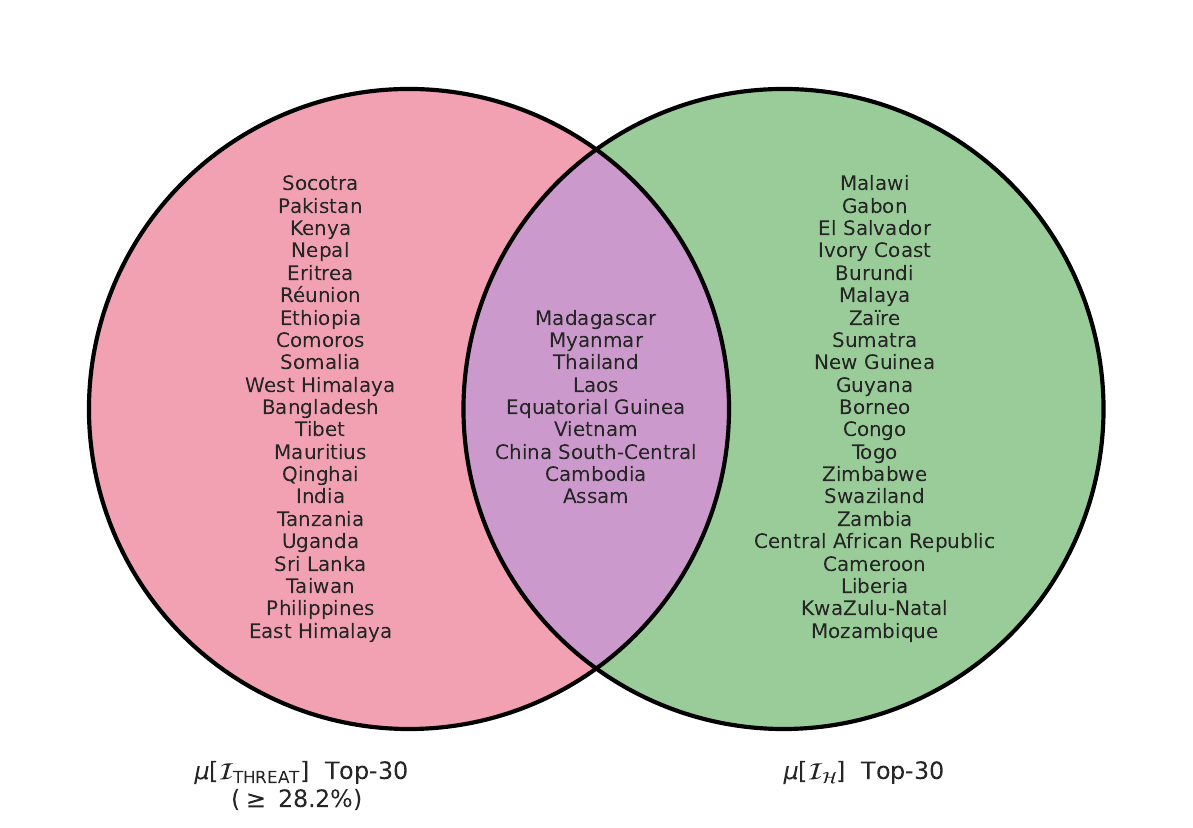}}
\centering
\textbf{(b)}\\
\vspace{0.25cm}
\begin{threeparttable}
\rowcolors{2}{gray!25}{white}

\begin{tabular}{lrr}
\headrow
\textbf{Continent     } &  \textbf{Coef.} &  $\rho$\textbf{-value} \\
\textbf{Africa        } &  0.206716 &  1.099494e-01 \\
\textbf{Asia-Temperate} &  0.499064 &  2.250013e-04 \\
\textbf{Asia-Tropical } &  0.028462 &  8.925731e-01 \\
\textbf{Australasia   } &  0.350000 &  3.558196e-01 \\
\textbf{Europe        } &  0.650844 &  5.441973e-06 \\
\textbf{North America } & -0.510027 &  5.537334e-06 \\
\textbf{Pacific       } & -0.300000 &  6.238377e-01 \\
\textbf{South America } &  0.294890 &  7.228730e-02 \\
\textbf{Global        } &  0.292719 &  2.553788e-07 \\
\hline
\end{tabular}

\end{threeparttable}

\caption{
(a) Venn diagram between the $\mu[\mathcal{I}_{\mathcal{H}}]$ and $\mu[\mathcal{I}_{\text{THREAT}}]$ top 30 countries.
With the exception of Madagascar and Equatorial Guinea, all the countries in the intersection are from South and South-East Asia.
El Salvador and Guyana are the only countries in the diagram that are not from Africa or Asia.\\
(b) Considering the same two variables, the Spearman correlation and the $\rho$-value for all countries of the same continent.
According to the $\rho$-values, the correlations are statistically significant only in Asia-Temperate, Europe and North America. Furthermore, looking at the scatter plot, we can see that the European and North American diversity ranges are limited.
Finally, it is only in Asia-Temperate that we observe a significant and positive correlation between threat levels and a wide diversity range.
}\label{si:venn_rho_pC}
\end{figure}
\clearpage

 \section{Further considerations on the modelling choices}
\label{si:model_data_consid}
When calibrating the species assemblage model, the choice of average error rate translates into a trade-off between model generalisation and over-prediction.
Indeed, imposing a lower error rate results in a lower probability threshold and larger predicted assemblages.
But are the newly retained species likely to be present, or are they unreasonably predicted?
This is a difficult question of model calibration, which we believe deserves more attention in future studies.\\

Over the validation set, an error is defined as the absence of the true label within the returned species assemblage.
What is effectively measured is the recall of the model, i.e. the proportion of relevant species that are successfully retained.
In contrast, precision, i.e. the proportion of retained species that are relevant to the test point, is not directly measured (and could not be without absence data).
However, precision is a positive function of the conditional probability threshold. Therefore, by maximising the threshold for a given target error rate, we also maximise precision.
Finally, a recent study confirms that as long as models are flexible enough and well fine-tuned to avoid overfitting, they make coherent predictions on spatially separated test data \citep{valavi_flexible_nodate}.

\section{Comparison with established indicators}
\label{si:Icomp}
The $\mu[\mathcal{I}_{\text{THREAT}}]$ top countries (i.e. countries with the highest average proportion of predicted threatened species) largely overlap with the countries identified in \citep{zizka_automated_nodate} as having the highest proportion of potentially threatened species.\\

\noindent The Red List of Ecosystems is a classification scheme of the risk of ecosystem collapse, with categories and an assessment process that mirror the IUCN Red List of Threatened Species \citep{keith_scientific_2013}. 
This promising indicator currently suffers from poor data coverage, with only 509 assessments registered as of April 2023 (\url{https://assessments.iucnrle.org/}, accessed on 26/04/23).
This comparison highlights the rare global consistency of our indicators.
We could also imagine ecosystem-level indicators that take into account the extinction risk of species assemblages in their construction.\\

\noindent Safeguarding ecosystems within the post-2020 global biodiversity framework requires robust indicators that capture different dimensions: area, integrity and risk of collapse \citep{nicholson2021scientific}.
Among the recommendations for selecting indicators, two are particularly relevant to our work:
\textit{4. greater testing and validation of indicators is required to understand their ecosystem relevance, reliability and ease of interpretation} and
\textit{5. the connection between global indicators and national or local policy and reporting needs strengthening}.
Our indicators meet recommendation five, but suffer from a lack of ground truthing to be confidently applied on the ground, as the fourth recommendation points out.\\

Protecting species for their evolutionary distinctiveness, combined with an IUCN threatened status, is the approach taken by EDGE (Evolutionary Distinct Globally Endangered, \cite{isaac2018use}).
While EDGE species must be officially listed as threatened by the IUCN in addition to having an above-average ED score (Evolutionary Distinctiveness), \cite{vitt2023global} developed a conservation prioritisation method based on ED and rarity as \textit{the number of occupied regions} or \textit{the area of occupancy}.
Here, the spatial ranges considered are compiled from the WCSP (World Checklist of Selected Plant Families) and GIFT (Global Inventory of Floras and Traits) databases.
Tropical Africa does not emerge as a clear priority hotspot as our indicators suggest.
However, they highlight the Neotropics and Southeast Asia as hotspots of richness, as does our Shannon index indicator.
They also identify islands as having particularly high numbers of rare and distinct species.
Interestingly, they point out that orchid ED is highly correlated with their richness (R² = 0.87).\\

Finally, the closest indicator to date from our work is the global extinction probability of terrestrial vascular plants \citep{verones_global_2022}.
In a given place, this indicator is high if many threatened species are known to occur there and/or if they have very small ranges.
However, we defend our kilometre-scale resolution and the novel way in which we calculate $\mathcal{I}_c$. This allows us to weight the contribution of species by their relative probability of occurrence.\\

\noindent Although the Shannon index measures not only community richness but also its evenness, global vascular plant richness maps such as \citep{cai_global_2022} are the closest available point of comparison.
Again, both the resolution and construction of our indicator differ from previous work.

\end{document}